\documentclass[preprint,12pt]{elsarticle}

\usepackage{amssymb}
\usepackage{amsmath,amssymb,amsfonts}
\usepackage{algorithmic}
\usepackage{graphicx}
\usepackage{textcomp}
\usepackage{multirow}
\usepackage{subfig}
\usepackage{listings}
\usepackage{booktabs}
\usepackage{soul}
\usepackage[hidelinks]{hyperref}
\usepackage{lineno}
\usepackage[table]{xcolor}
\usepackage[utf8]{inputenc}
\usepackage[acronym]{glossaries}

\newcommand{\ctb}{\color{blue}}     
\newcommand{\ctr}{\color{red}}      

\makeglossaries

\newacronym{sota}{SOTA}{State-Of-The-Art}
\newacronym{fscil}{FSCIL}{Few-Shot Class-Incremental Learning}
\newacronym{cil}{CIL}{Class Incremental Learning}
\newacronym{robusta}{ROBUSTA}{Robust Transformer Approach}
\newacronym[plural=CNNs]{cnn}{CNN}{Convolutional Neural Networks}
\newacronym[plural=CCTs]{cct}{CCT}{Compact Convolutional Transformer}
\newacronym[plural=MLPs]{mlp}{MLP}{Multi-Layer Perceptron}
\newacronym[plural=IDs]{id}{ID}{identifiers}
\newacronym{fsl}{FSL}{Few-Shot Learning}
\newacronym{cf}{CF}{Catastrophic Forgetting}
\newacronym{cl}{CL}{Continual Learning}
\newacronym{mhsa}{MHSA}{Multi-Head Self-Attention}
\newacronym{ng}{NG}{Neural Gas}
\newacronym{gat}{GAT}{Graph Attention Network}
\newacronym{mi}{MI}{Mini-ImageNet}
\newacronym{sgd}{SGD}{Stochastic Gradient Descent}
\newacronym{mse}{MSE}{Mean Squared Error}
\newacronym{bn}{BatchNorm}{Batch Normalization}

\journal{Information Sciences}

\begin{document}

\begin{frontmatter}



\title{Few-Shot Class Incremental Learning via Robust Transformer Approach}



\author[UniSA]{Naeem Paeedeh\fnref{fn1}}
\ead{naeem.paeedeh@mymail.unisa.edu.au}

\author[UniSA]{Mahardhika Pratama\fnref{fn1}}
\ead{dhika.pratama@unisa.edu.au}

\author[UGM]{Sunu Wibirama}
\ead{sunu@ugm.ac.id}

\author[UniSA]{Wolfgang Mayer}
\ead{wolfgang.mayer@unisa.edu.au}

\author[UniSA]{Zehong Cao}
\ead{jimmy.cao@unisa.edu.au}

\author[UniSA,SRI]{Ryszard Kowalczyk}
\ead{ryszard.kowalczyk@unisa.edu.au}

\fntext[fn1]{N. Paeedeh and M. Pratama share equal contributions.}

\affiliation[UniSA]{organization={STEM, University of South Australia},
            city={Adelaide},
            state={South Australia},
            country={Australia}
            }

\affiliation[SRI]{organization={Systems Research Institute, Polish Academy of Sciences},
    country={Poland}
}
\affiliation[UGM]{organization={Department of Electrical and Information Engineering, University of Gadjah Mada}, country={Indonesia}}

\begin{abstract}
\Gls{fscil} presents an extension of the \Gls{cil} problem where a model is faced with the problem of data scarcity while addressing the \Gls{cf} problem. This problem remains an open problem because all recent works are built upon the \Glspl{cnn} performing sub-optimally compared to the transformer approaches. Our paper presents \Gls{robusta} built upon the \Gls{cct}. The issue of overfitting due to few samples is overcome with the notion of the stochastic classifier, where the classifier's weights are sampled from a distribution with mean and variance vectors, thus increasing the likelihood of correct classifications, and the batch-norm layer to stabilize the training process. The issue of \Gls{cf} is dealt with the idea of delta parameters, small task-specific trainable parameters while keeping the backbone networks frozen. A non-parametric approach is developed to infer the delta parameters for the model's predictions. The prototype rectification approach is applied to avoid biased prototype calculations due to the issue of data scarcity. The advantage of \Gls{robusta} is demonstrated through a series of experiments in the benchmark problems where it is capable of outperforming prior arts with big margins without any data augmentation protocols. 
\end{abstract}



\begin{keyword}
\acrlong{cl} \sep \acrlong{cil} \sep \acrlong{fscil} \sep \acrlong{fsl}



\end{keyword}

\end{frontmatter}


\section{Introduction}
%
%
%
%
\Gls{fscil}~\cite{Tao2020FewShotCL} forms an extension of \Gls{cl}~\cite{Parisi2019ContinualLL} where the underlying goal is to cope with the problem of sparse data in \Gls{cl}. That is, a model is presented with a base task containing large samples, and a sequence of few-shot learning tasks formulated in the N-way K-shot setting, i.e., a task is composed of $N$ classes with $K$ samples per class in which $K$ is usually small, e.g., $K=1$ or $K=5$. Consequently, a \Gls{cl} agent has to deal with the over-fitting problem and the \Gls{cf} problem simultaneously when learning new tasks. Note that the \Gls{cf} problem exists because old parameters are overwritten when learning new tasks.

The \Gls{fscil} problem is pioneered in \cite{Tao2020FewShotCL}, where a solution is proposed based on the \Gls{ng} approach. \cite{Mazumder2021FewShotLL} puts forward the notion of session trainable parameters to overcome the \Gls{fscil} while \cite{Chen2021IncrementalFL} develops its solution using the vector quantization method. \cite{Zhang2021FewShotIL} finds the performance of the base task as a key success of the \Gls{fscil} problem and proposes the \Gls{gat} to adapt the classifier's weights. \cite{Shi2021OvercomingCF} proposes the flat learning concept where the flat regions are found in the base learning tasks and maintained when learning sequences of few-shot learning tasks to address the \Gls{cf} problem. \cite{Kalla2022S3CSS} makes use of the stochastic classifier idea \cite{Lu2020StochasticCF} coupled with the self-supervised learning strategy. Nevertheless, the performance of these works heavily depends on the base learning task, and performance degradation is expected when the classes of the base learning task shrink. In addition, all these works are built upon the convolutional structures which perform sub-optimally compared to the transformer backbone.

This paper proposes \Gls{robusta} for \Gls{fscil} constructed under the \Gls{cct} backbone \cite{Hassani2021EscapingTB}. The \Gls{cct} allows a compact structure due to the use of a convolutional tokenizer in terms of sequence pooling and convolutional layer, downsizing the patch size, and removing the class token and positional embedding.  \Gls{robusta} is developed from the notion of stochastic classifier \cite{Lu2020StochasticCF} to cope with the over-fitting issue, where the classifier weights are represented as the mean and variance vector. That is, classifier weights are sampled from a distribution. This assures the presence of infinite classifiers, thus supporting correct classifications. In addition, the batch norm layer is integrated in lieu of the layer norm as per the conventional transformer to stabilize the training process \cite{Yao2021LeveragingBN}. The concept of delta parameters is developed where small task-specific trainable parameters are incorporated while freezing the backbone network \cite{Wang2021LearningTP, Wang2022DualPromptCP, Wang2023RehearsalfreeCL}. To avoid any dependencies on the task \Glspl{id}, a non-parametric approach is put forward to infer the delta parameters for the model's predictions. Last but not least, the concept of prototype rectification is implemented to prevent biased prototype calculations. That is, it overcomes the problem of intra-class bias due to the issue of data scarcity \cite{Liu2019PrototypeRF, Xue2020OneShotIC}.

The base training phase of \Gls{robusta} is set to minimize the supervised learning loss and the self-supervised learning loss via DINO \cite{Caron2021EmergingPI}. The DINO technique is chosen here because it is specifically designed for the transformer backbone. The supervised learning method aims to master the base classes while the generalization beyond the base classes is availed by the self-supervised learning phase via the DINO technique, thus avoiding supervision collapses \cite{Doersch2020CrossTransformersSF}. The concept of delta parameters is incorporated and follows the parameter isolation strategy. That is, small trainable parameters are pre-pended to the \Gls{mhsa} layer for every task. Only the delta parameters are learned while leaving other parameters fixed. The non-parametric approach is developed to select the delta parameters for inferences \cite{Wang2021LearningTP, Wang2022DualPromptCP, Wang2023RehearsalfreeCL}. Note that the use of memories violates the \Gls{fscil} spirit because it might lead to replaying all samples due to the few sample constraints. All of these are performed under the \Gls{cct} backbone with the stochastic classifier, making it possible to create arbitrary numbers of classifiers underpinning improved classifications and the integration of batch norm layer expediting the model's convergence. The idea of prototype rectification is proposed, where a prediction network is trained to predict prototypes. The predicted prototypes and the original prototypes are aggregated to produce the correct prototypes. The following explains the roles of each learning component of \Gls{robusta} to address the major problems of \Gls{fscil}.
\begin{itemize}
    \item \textbf{Over-fitting Problem}: the problem of over-fitting exists in the \Gls{fscil} problem because of very few samples in the few-shot tasks. This problem is addressed in \Gls{robusta} by the use of delta parameters and a stochastic classifier. The use of delta parameters fixes the backbone network leaving very small trainable parameters mitigating the risks of over-fitting. Besides, the use of stochastic classifiers allows flexibility in the output space because the classifier weights are sampled from a distribution, leading to the presence of arbitrary numbers of classifiers and increasing the likelihood of correct classifications. 
    \item \textbf{Catastrophic Forgetting Problem}: the problem of catastrophic forgetting is seen because of a sequence of few-shot learning tasks causing old parameters to be over-written when learning new tasks. This issue is addressed specifically by freezing the backbone network. That is, the backbone network is frozen after it is trained by the supervised learning phase and the self-supervised learning phase in the base task. The concept of task-specific delta parameters is introduced to learn the incremental tasks, while the non-parametric approach is applied to infer such delta parameters during the testing phase. 
    \item \textbf{Intra-Class Bias}: the problem of intra-class bias occurs in the prototype estimation phase because of very few samples in the few-shot learning phase. That is, very few samples do not describe the true class distributions. This problem is overcome by a prototype rectification strategy via a prediction network. The prediction network is trained to estimate the actual prototype, thus compensating for the intra-class bias problem.
\end{itemize} 

Our paper conveys at least four contributions: 
\begin{enumerate}
    \item \textcolor{red}{We propose the concept of the stochastic classifier for \Gls{fscil}. The key difference with prior works in \cite{Lu2020StochasticCF, Kalla2022S3CSS} lies in which such stochastic classifier is integrated under the transformer backbone};
    \item \textcolor{green}{We propose the batch norm layer to stabilize the learning process instead of the layer norm because we handle the visual problems rather than texts \cite{Yao2021LeveragingBN}. Note that the original transformer layer makes use of the layer norm concept, and direct replacement of layer norm with batch norm might cause instability in the training process};
    \item \textcolor{red}{The concept of delta parameters, along with the non-parametric parameter identification strategy, are proposed to combat the issue of \Gls{cf} with small additional expenditures. To the best of our knowledge, we are the first to adopt this concept for the \Gls{fscil} problems  \cite{Wang2021LearningTP, Wang2022DualPromptCP, Wang2023RehearsalfreeCL} where the issues of overfitting and \Gls{cf} problem are to be handled simultaneously};
    \item \textcolor{red}{We propose the idea of prototype rectifications to prevent biased prototype calculations due to the intra-class bias as a result of data scarcity. This is enabled by a prediction network trained to infer correct prototypes. Our approach differs from \cite{Xue2020OneShotIC, Liu2019PrototypeRF} because such a concept is embedded in the \Gls{fscil} problem rather than the conventional few-shot learning}; 
\end{enumerate}
Our rigorous experiments in the benchmark problems demonstrate the advantage of \Gls{robusta}, where it outperforms prior arts with $1-9\%$ margins without any data augmentation strategies. Our source codes are made publicly available in \url{https://github.com/Naeem-Paeedeh/ROBUSTA} for reproducibility and convenient further study. Table \ref{table:acronyms} and Table \ref{table:notations} list the acronyms and notations used in this paper, respectively.

\section{Related Works}
\subsection{Continual Learning}
The \Gls{cl} is a research area of growing interest where the goal is to develop a lifelong learning agent that improves its intelligence over time \cite{Parisi2019ContinualLL}. The main challenge is to overcome the \Gls{cf} problem of previously observed knowledge because old parameters are erased when learning new tasks. Three approaches in the literature exist to address the \Gls{cf} problem \cite{Parisi2019ContinualLL}: regularization-based, architecture-based, and memory-based approaches. The regularization-based approach \cite{Kirkpatrick2017OvercomingCF, Zenke2017ContinualLT, Aljundi2018MemoryAS, Cha2021CPRCR, Paik2020OvercomingCF, Li2016LearningWF, Mao2021ContinualLV} relies on a regularization term or a learning rate adjustment preventing important parameters of old tasks from significant deviations. This approach is easy to implement and relatively fast. Still, its performances are usually poor in the \Gls{cil} setting because of the absence of any task \Glspl{id}. In addition, it is hindered by the fact that an overlapping region of all tasks is relatively difficult to find, notably for large-scale \Gls{cl} problems. The architecture-based approach \cite{Rusu2016ProgressiveNN, Yoon2018LifelongLW, Li2019LearnTG, Xu2021AdaptivePC, Ashfahani2022UnsupervisedCL, Pratama2021UnsupervisedCL, Zoph2017NeuralAS, Rakaraddi2022ReinforcedCL} takes different routes where a new task is handled by introducing extra network capacities while isolating previous network parameters. As with the regularization-based approach, this approach depends on the task \Glspl{id} and is computationally prohibitive, especially when involving neural architecture searches to adjust network structures toward new tasks. The memory-based approach \cite{Rebuffi2017iCaRLIC, Castro2018EndtoEndIL, Hou2019LearningAU, Chaudhry2021UsingHT, Chaudhry2019EfficientLL, Buzzega2020DarkEF, Dam2022ScalableAO, VinciusdeCarvalho2022ClassIncrementalLV,Masum2023AssessorGuidedLF} relies on a small episodic memory for experience replay. Although this method often performs better than the previous two approaches, it imposes an extra memory burden, i.e., thousands of samples must be stored in the memory. This approach is also incompatible with the \Gls{fscil} because it becomes equivalent to the retraining approach, violating the spirit of \Gls{cl} due to the sample scarcity constraints. Our approach, \Gls{robusta}, relies on the architecture-based approach where small trainable parameters, namely delta parameters, are inserted in every incremental task, leaving other parameters fixed to cope with the \Gls{cf} problem. 

\subsection{Few-Shot Class-Incremental Learning}
The \Gls{fscil} is put forward in \cite{Tao2020FewShotCL} as an extension of general \Gls{cl} problems where it aims to cope with the issue of data scarcity in continual environments. \cite{Tao2020FewShotCL} proposes the \Gls{ng} solution to address such challenges. \cite{Mazumder2021FewShotLL} tries different approaches where the session-trainable parameters are put forward to reduce the \Gls{cf} and overfitting problems simultaneously. The vector quantization approach is developed in \cite{Chen2021IncrementalFL} while \cite{Zhang2021FewShotIL} introduces the use of \Gls{gat}. \cite{Shi2021OvercomingCF} devises the flat-minima solution to avoid the \Gls{cf} problems, while \cite{Kalla2022S3CSS} relies on the idea of stochastic classifier \cite{Lu2020StochasticCF} coupled with the self-supervised learning strategy. All of these emphasize the importance of the base task for the \Gls{fscil} because it comprises many classes. Their performances deteriorate in the case of small base tasks involving few classes in the base task. It is worth mentioning that \cite{Shi2021OvercomingCF} utilizes the episodic memory for experience replay, which does not fit the \Gls{fscil} problems, and \cite{Kalla2022S3CSS} makes use of the data augmentation strategy simplifying the \Gls{fscil} problem to a great extent.

\section{Problem Definition}
A \Gls{fscil} problem is defined as a learning problem of sequentially arriving few-shot learning tasks $\mathcal{T}_1,\mathcal{T}_2,...,\mathcal{T}_K$, $k\in K$ where each few-shot learning task possesses tuples of data points $\mathcal{T}_{k}=\{(x_i,y_i)\}_{i=1}^{N_k}$ drawn from the same domain $\mathcal{X}\times\mathcal{Y}\in\mathcal{D}$. $x\in\mathcal{X}, y\in\mathcal{Y}$ respectively denote the input image and its corresponding label. The \Gls{fscil} problem features the data scarcity problem where each task $\mathcal{T}_k$ is configured in the N-way K-shot setting, i.e., each task consists of N classes where each class only carries K samples, thus leading to the overfitting issue. A base task $\mathcal{T}_0=\{(x_i,y_i)\}_{i=1}^{N_0}$ is given where there exist moderate samples $N_0>>N_k$. Suppose that $L_k, L_{k^{'}}$ stand for the label sets of the $k-th$ and $k^{'}-th$ tasks respectively, the \Gls{fscil} follows the conventional \Gls{cil} problem \cite{Ven2019ThreeSF} where each task possesses disjoint class labels with the absence of any task \Glspl{id} $\forall k, k^{'} L_k\cap L_{k^{'}}=\emptyset$. A task $\mathcal{T}_k$ is discarded once observed and thus opens the risk of \Gls{cf} problem. Due to the few sample constraints, the use of memory is infeasible, while the use of data augmentation strategies results in over-simplifications of the \Gls{fscil} problems.\par

\begin{figure}
    \centering
    \includegraphics[scale=0.8]{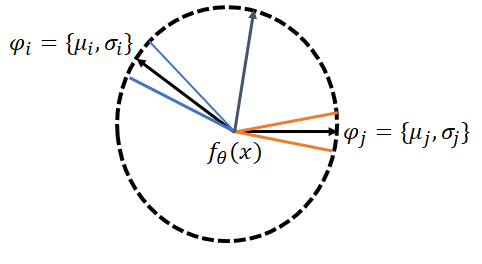}
    \caption{Illustration of Stochastic Classifier}
    \label{fig:stochastic_classifier}
\end{figure}

\begin{figure}[h]
  \centering
  \includegraphics[width=\linewidth]{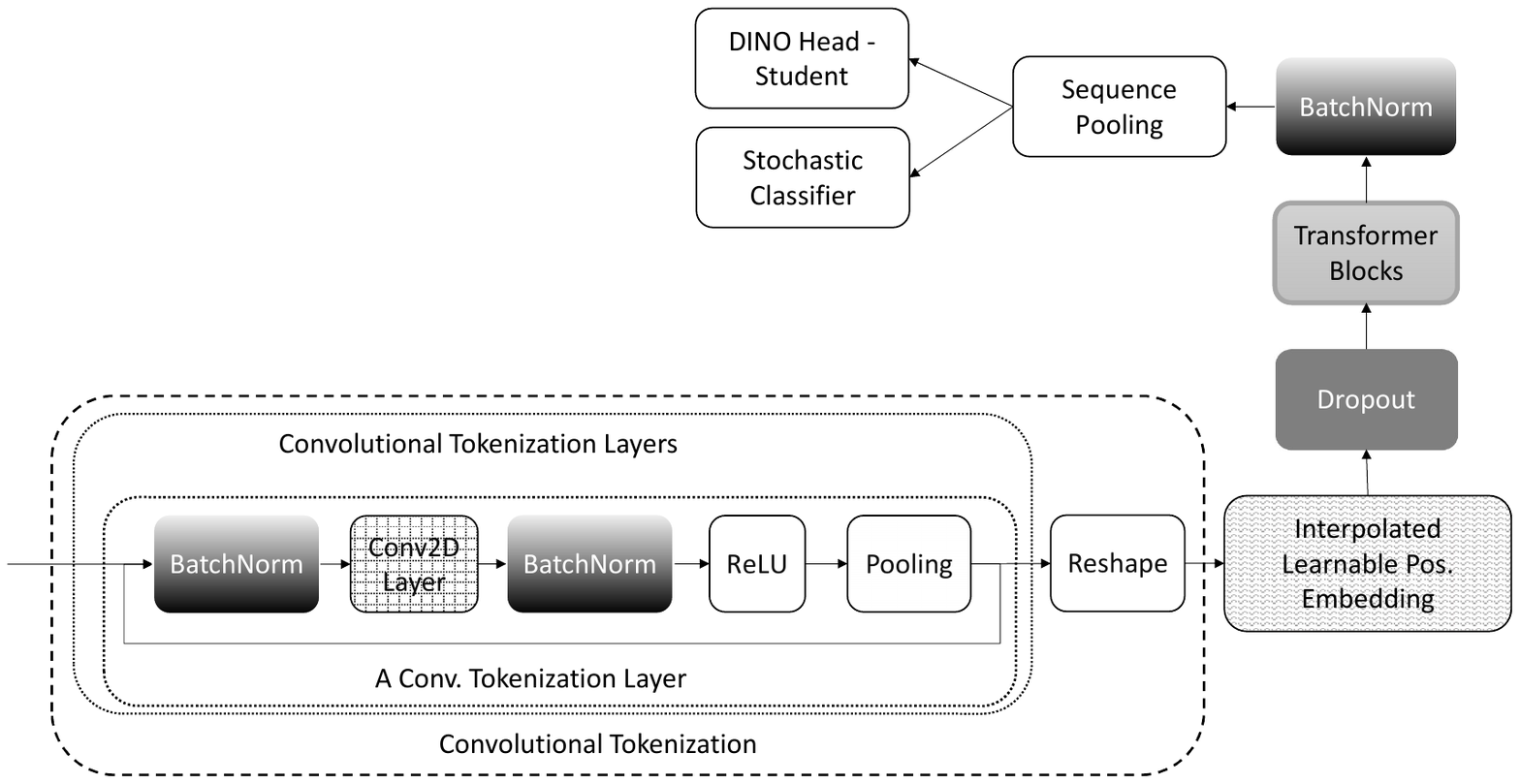}
  \caption{Overview of the student model. We replace the LayerNorm layers in the \Gls{cct} model with \Gls{bn} layers.}
  \label{fig: student}
\end{figure}

\begin{figure}[h]
  \centering
  \includegraphics[width=\linewidth]{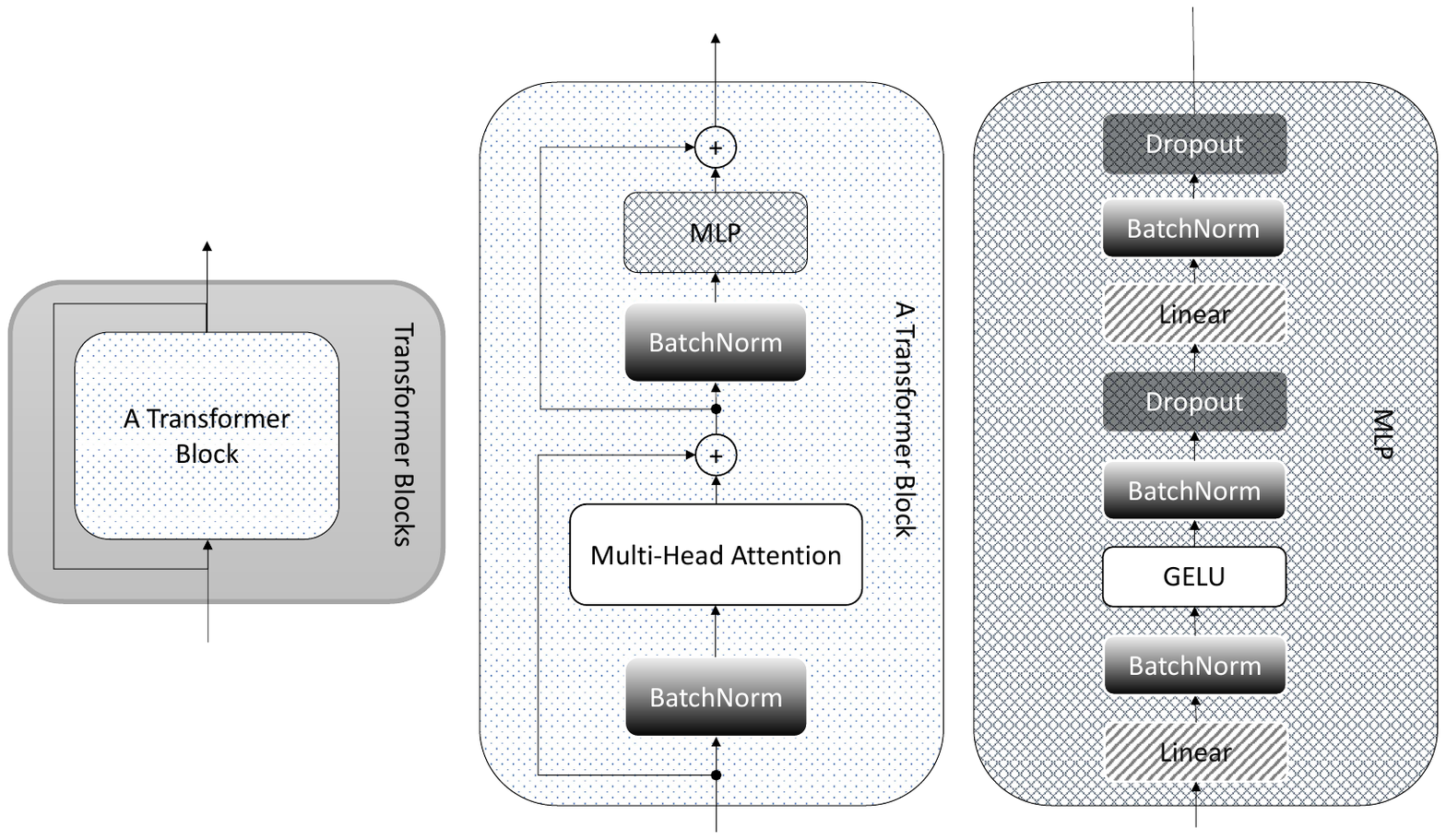}
  \caption{The blocks of transformers. We replace the LayerNorm layers in the transformer blocks. Moreover, we add a \Gls{bn} layer between the two linear layers of the \Gls{mlp}.}
  \label{fig: architecture}
\end{figure}

\begin{table*}[!tb]
    \centering
    \caption{A list of the key acronyms}
    \label{table:acronyms}
    \resizebox{\textwidth}{!}{
        \begin{tabular}{ |>{\color{black}}l>{\color{black}}l|>{\color{black}}l>{\color{black}}l| }
            \toprule
            \multicolumn{4}{|c|}{Nomenclature} \\
            \bottomrule
            Full Name & Abbreviation & Full Name & Abbreviation \\
            \bottomrule
            \acrlong{fscil}   & \acrshort{fscil}   & \acrlong{cf} & \acrshort{cf} \\
            \acrlong{cil}     & \acrshort{cil}     & \acrlong{cl} & \acrshort{cl} \\
            \acrlong{robusta} & \acrshort{robusta} & \acrlong{mhsa} & \acrshort{mhsa} \\
            \acrlong{cnn}     & \acrshort{cnn}     & \acrlong{ng} & \acrshort{ng} \\
            \acrlong{cct}     & \acrshort{cct}     & \acrlong{cl} & \acrshort{cl} \\
            \acrlong{id}      & \acrshort{id}      & \acrlong{mlp} & \acrshort{mlp}  \\
            \acrlong{mse}      & \acrshort{mse}    & \acrlong{sgd} & \acrshort{sgd}   \\
            \acrlong{gat}      & \acrshort{gat}    & \acrlong{bn} & \acrshort{bn}    \\
            \acrlong{mi}      & \acrshort{mi}    & &    \\
            \hline
        \end{tabular}
    }
\end{table*}


\begin{table*}[!tb]
    \centering
    \caption{A list of the key notations}
    \label{table:notations}
    \resizebox{\textwidth}{!}{
        \begin{tabular}{ |>{\color{black}}l|>{\color{black}}l| }
            \toprule
            Mathematical symbol & Explanation\\
            \bottomrule
            $k\in K$ & Number of class-incremental tasks \\
            $N_k$ & Number of samples in a task \\
            $\mathcal{X}\times\mathcal{Y}\in\mathcal{D}$ & The domain of the inputs \\
            $x\in\mathcal{X}$ & Input image\\
            $y\in\mathcal{Y}$ & Label \\
            $\mathcal{T}_k=\{(x_i,y_i)\}_{i=1}^{N_k}$ & The samples set of task $k$ \\
            $\mathcal{T}_1,\mathcal{T}_2,...,\mathcal{T}_K$, $k\in K$ & Few-shot learning tasks \\
            $L_k$ & Labels set for the $k-th$ task \\
            $n$ & Sequence length \\
            $d$ & The dimension of the head (embeddings) \\
            $X\in\Re^{n\times d}$ & Image patches \\
            $H$ & Number of heads \\
            $h$ & Head index \\
            $Q$ & Query in the self-attention components \\
            $K$ & Key in the self-attention components \\
            $V$ & Value in the self-attention components \\
            $h_Q$ & Input query \\
            $h_K$ & Input key \\
            $h_V$ & Input value \\
            $MHSA(Q, K, V)$ & Multi-Head Self-Attention\\
            $\theta^{O}$ & Linear projection, $\theta_h^{Q},\theta_h^{K},\theta_h^{V}\in\Re^{d\times d_k}$ \\
            $\theta_h^{Q},\theta_h^{K},\theta_h^{V}\in\Re^{d\times d_k}$ & Linear projections for the query, key, value in the self-attention \\
            $\Psi_h$ & Self-attention matrix, $\Psi_h\in\Re^{n\times n}$ \\
            $\Psi_h^{i,j}$ & Attention score that from token $i$ to token $j$ \\
            $\sigma(.)$ & Softmax activation function \\
            $FFN(.)$ & Feed-forward module \\
            $\theta_1\in\Re^{d\times d'}, \theta_2\in\Re^{d'\times d}$ & Projection matrices in the feed-forward module \\
            $\theta_{BN}\in\Re^{d\times d}$ & The parameters of the \Gls{bn} layer \\
            $z\in\Re^{D}$ & Feature space \\
            $f_{\theta}(x):\mathcal{X}\rightarrow \mathcal{Z}$ & Feature extractor that maps the input to the feature space $z\in\Re^{D}$ \\
            $g_{\phi}(z):\mathcal{Z}\rightarrow \mathcal{Y}$ & The classifier that maps a feature vector to a label score \\
            $\mu,\sigma$ & Mean and variance of the Gaussian distribution for the stochastic classifier \\
            $\phi_i=\{\mu_i,\sigma_i\}$ & The weights of the stochastic classifier \\
            $\eta$ & Temperature parameter of the stochastic classifier that controls the smoothness of the output distribution \\
            $l(.)$ & Cross-entropy loss function \\
            $f_{W_{t}}(.)$ & The teacher network \\
            $f_{W_{s}}(.)$ & The student network \\
            $x_1^g$ and $x_2^g$ & The global-view crops of the image for the teacher network \\
            $V$ & The set of all global and local crops for the student network \\
            $\hat{.}$ & $L_2$ normalization of a vector \\
            $p_k$ & Delta-parameters for the task $k$\\
            $L_p$ & The length of the prefixes \\
            $p_K, p_V \in \Re^{(L_p/2\times d)}$ & Prefixes to be prepended to the inputs of the key and value in the self-attention \\
            $N_c$ & Number of training samples of the $c-th$ class \\
            $u_t^c$ & The average of the outputs for the $c-th$ class of task $t$ \\ 
            $A$ & The covariance matrix \\ 
            $c^{*}$ & The selected class at the inference phase \\
            $B_{intra}$ & The intra-class bias \\
            $P_{\zeta}(.):\mathcal{Z}\rightarrow\mathcal{\mu}$ & The prediction network that maps the embeddings to the prototype space \\
            $\lambda$ & The outliers \\
            $R(.)$ & The prediction net as a function that refines the prototypes \\
            $\mu_c$ & The refined prototypes  \\
            \hline
        \end{tabular}
    }
\end{table*}


\section{Method}
\subsection{Network Architecture}
\Gls{robusta} is built upon the \Gls{cct} backbone \cite{Hassani2021EscapingTB} in which the convolutional tokenization is applied without any positional embedding and class tokens, thus scaling well for limited data situations. However, the original \Gls{cct} implements a conventional transformer encoder with the layer normalization technique. The layer normalization approach inherits the original transformer architectures for texts because it can fit any sequence length. The \Gls{bn} approach is more effective than layer normalization since the image classification problems are handled here where the input size is fixed. It allows faster convergence than models with layer normalization. Our innovation here is to apply the \Gls{bn} layer in lieu of the layer normalization method under the \Gls{cct} backbone. As investigated by \cite{Yao2021LeveragingBN}, simple replacements of layer normalization with \Gls{bn} lead to frequent crashes in convergence due to un-normalized feed-forward network blocks. We choose to insert the \Gls{bn} layer between two linear layers of the feed-forward network blocks of the \Gls{cct}. 

Given a sequence of image patches $X\in\Re^{n\times d}$ where $n,d$ respectively denote the sequence length and the vector dimension, the input sequence is processed by a stack of encoder layers comprising the \Gls{mhsa} module and the feed-forward layer \cite{Xue2022MetaattentionFV}. The \Gls{mhsa} module consists of several heads with their own query $Q_h$, key $K_h$, and value $V_h$, whose outputs are concatenated and linearly projected as follows:
\begin{equation}
    MHSA(Q, K, V)=Concat(head_1,...,head_H)\theta^{O}
\end{equation}\label{MHSA}
where $H$ is the number of heads. $head_h\in\Re^{n\times d_k}$ where $d_k=d/H$ and $\theta_h^{O}\in\Re^{(d_k\times d)}$. The output of attention head $head_h$ is formalized as follows:
\begin{equation}\label{self-attention}
    head_h=\Psi_h V_h =\sigma(A_h)V_h=\sigma(\frac{Q_h K_h^T}{\sqrt{d_k}})V_h
\end{equation}
where the query, key, value embedding $Q_h,K_h,V_h$ are resulted from the linear projections of the input sequence $X$ and the projectors $\theta_h^{Q},\theta_h^{K},\theta_h^{V}\in\Re^{d\times d_k}$.
\begin{equation}
    \begin{split}
        Q_h=X\theta_h^{Q}\\K_h=X\theta_h^{K}\\V_h=X\theta_h^{V}
    \end{split}
\end{equation}
where $\theta_h^{Q},\theta_h^{K},\theta_h^{V}\in\Re^{d\times d_k}$ are learnable parameters. $\sigma(.)$ is the softmax activation function and $\Psi_h\in\Re^{n\times n}$ is a self-attention matrix indicating the similarity between the query and key pairs. $\Psi_h^{i,j}$ is the attention score that token $i$ pays to token $j$. There exist two linear layers with the GELU activation function $\psi(.)$ and the \Gls{bn} in the feed-forward module:
\begin{equation}\label{fully-connected}
    FFN(X;\theta_1,\theta_2,\theta_{BN})=\psi((X \theta_{BN}) \theta_1)\theta_2
\end{equation}
where $\theta_1\in\Re^{d\times d'}, \theta_2\in\Re^{d'\times d}$ are projection matrices while $\theta_{BN}\in\Re^{d\times d}$ is the parameters of the \Gls{bn}. Fig. \ref{fig: student} and \ref{fig: architecture} visualize the network structure of \Gls{robusta}.

Given a feature extractor $f_{\theta}(x):\mathcal{X}\rightarrow \mathcal{Z}$ embedding an input image $x$ to the feature space $z\in\Re^{D}$ and a classifier $g_{\phi}(z):\mathcal{Z}\rightarrow \mathcal{Y}$ converting a feature vector to a label score, the cosine similarity is used to measure the label score for a feature of interest $\langle\Hat{f}_{\theta}(x),\Hat{\phi}_i\rangle$ where $\hat{u}=u/||u||_2$ denotes the $L_2$ normalized vector and $\phi_{i}$ stands for the classifier weight. The stochastic classifier is portrayed in Fig. \ref{fig:stochastic_classifier} where the classifier weight is defined as a distribution $\phi_i=\{\mu_i,\sigma_i\}$. $\mu,\sigma$ stands for the mean and variance of the Gaussian distribution. This strategy allows the creation of an arbitrary number of classifiers, increasing the chance of correct classifications. To allow end-to-end training, the following re-parameterization trick is applied:
\begin{equation}\label{eq:reparameterization}
\langle\hat{\phi},\hat{f}_{\theta}(x)\rangle,(\phi=\mu+\mathcal{N}(0,1)\odot\sigma),
\end{equation}
where $\mu,\sigma$ are learned during the training process while $\odot$ is an element-wise multiplication operator. The inference process of \Gls{robusta} is formalized:
\begin{equation}\label{output}   P(Y=m|x)=\frac{\exp{(\eta\langle\hat{\phi}_{m},\hat{f}(x)}\rangle)}{\sum_{m=1}^{M}\exp{(\eta\langle\hat{\phi}_{m},\hat{f}(x)}\rangle)}
\end{equation}
where $M$ is the number of classes and $\eta$ is the temperature that controls the smoothness of the output distribution. The final output is calculated as $\arg\max_{1\leq m\leq M}P(Y=m|x)$. We adopt the same approach as \cite{Kalla2022S3CSS} where the means are initialized as the class prototypes. Compared to the conventional ViT \cite{Dosovitskiy2020AnII}, the network architecture of \Gls{robusta} features three key differences 
\begin{itemize}
    \item Our architecture adopts the convolutional tokenizer to produce the patch embedding. It reduces the patch size and removes the class token and positional embedding. 
    \item Our architecture incorporates the batch norm rather than the layer norm to stabilize the learning process. This modification is suited to our context focusing on the image classification rather than texts. 
    \item Our architecture develops the concept of the stochastic classifier. It goes one step ahead of conventional prototypes where the classifier weights are sampled from a distribution, increasing the likelihood of correct classifications, thus being well suited to the data scarcity problem. 
\end{itemize}

\subsection{Base Learning Task $\mathcal{T}_{k=1}$}
The base learning phase of \Gls{robusta} consists of two joint learning objectives: the supervised learning phase and the self-supervised learning phase. The supervised learning phase is meant to master the base classes, while the self-supervised learning phase extracts general features, making it possible to generalize beyond the base classes. The supervised learning phase applies the conventional cross-entropy loss function while the self-supervised learning phase implements the DINO technique \cite{Caron2021EmergingPI} based on the knowledge distillation technique between the teacher network $f_{W_{t}}(.)$ and the student network $f_{W_{s}}(.)$. As with the DINO approach, the multi-crop strategy is implemented to produce different views of an image.

\begin{equation}\label{baseloss}
\mathcal{L}_{k=1}=\mathbb{E}_{(x,y)\backsim\mathcal{D}_1}[l(y,\hat{y})+\sum_{x\in\{x_1^{g},x_2^{g}\}}\sum_{\tilde{x}\in V}l(f_{W_t}(x),f_{W_s}(\tilde{x}))]
\end{equation}
where $l(.)$ stands for the cross-entropy loss function and $x\neq\tilde{x}$. Note that the global views of the image $x^{g}$ are applied to the teacher only while all crops are fed to the student network. The same architecture is configured for both the teacher and student networks. \eqref{baseloss} is used to update the student network only, i.e., the stop gradient is implemented for the teacher network. The teacher network is momentum updated and passed to the next tasks $\mathcal{T}_{k>1}$ while its outputs are centered and sharpened to prevent the model's collapses. The cross-entropy loss function is calculated by using the stochastic classifier $\langle\hat{f}_{\theta}(x),\hat{\phi}_i\rangle$ as the classification head with the cosine similarity as per \eqref{output}. That is, $(\mu,\sigma)$ are updated along with other network parameters in the end-to-end fashion. The alternate optimization strategy is applied to solve \eqref{baseloss} where a model is tuned using the DINO loss function until the completion, followed by the cross entropy optimization procedure. 
\subsection{Few-Shot Learning Task $\mathcal{T}_{k>1}$}
\subsubsection{Delta Parameters}
In the few-shot learning phases, a model encounters the \Gls{cf} problem because a model receives a sequence of tasks. That is, previously valid parameters are overwritten when learning new tasks, causing dramatic performance drops for previous tasks. One possible solution is via the use of different parameters for different tasks \cite{Wang2023RehearsalfreeCL}, but this naive solution leads to prohibitive storage costs. We approach this problem using the notion of prefix tuning \cite{Wang2022DualPromptCP, Wang2021LearningTP} where a small number of trainable parameters are inserted in the \Gls{mhsa} layer, leaving the remainder of parameters fixed to prevent interference. 

For the $k-th$ task, a set of task-specific parameters, namely delta parameters $p_k$, are prepended to the input of \Gls{mhsa} block at every layer. Let the input to the \Gls{mhsa} layer be $h\in\Re^{L\times D}$, and we further label the input query, key, and values for the \Gls{mhsa} layer to be $h_Q,h_K,h_V$ respectively. The prefix tuning splits $p$ into $p_K, p_V\in\Re^{(L_p/2\times d)}$ and prepends them to $h_K$ and $h_V$ respectively, while keeping $h_Q$ intact. 
\begin{equation}
    f_{Pre-T}(p,h)=MHSA(h_{Q},[p_k;h_K],[p_v;h_V])
\end{equation}
where $[;]$ is the concatenation operation along the sequence length dimension. The prefix tuning does not alter the output dimension, which remains the same as the input dimension $\Re^{L\times D}$. We only train the delta parameters and the classification head during the training process. This leads to significant reductions of network parameters to be trained, e.g., only $0.27\%$ of the total network parameters \cite{Wang2023RehearsalfreeCL}.   
\subsubsection{Task Inference}
Because of the use of task-specific delta parameters, a model is required to infer the delta parameters to avoid dependence on the task \Glspl{id} during the testing phase. One possible approach is to select the one returning the highest confidence score, but such an approach is often over-confident for unrelated tasks \cite{Wang2023RehearsalfreeCL}. Another approach is via the nearest neighbor strategy, where a test sample is compared to that of the training sample. Such an approach imposes the use of memory, which is impossible to apply in the realm of the \Gls{fscil} problems.  

Our task inference strategy is via a non-parametric approach where a test sample is compared with full distributions of the task's training samples \cite{Wang2023RehearsalfreeCL}. That is, we model the distribution of training samples using the Gaussian distribution. Let $h=g_{\phi}(f_{\theta}(x))$ be the last transformer block's outputs, the mean $\mu_k^{c}$ and covariance $A_k$ of the Gaussian distributions are estimated using the maximum likelihood estimation as follows:
\begin{equation}
    \mu_{k}^{c}=\frac{1}{N_c}\sum_{y_i=c}h(x_i)
\end{equation}
\begin{equation}
    A_{k}=\frac{1}{N_c}\sum_{c}\sum_{y_i=c}(h(x_i)-\mu_{k}^{c})(h(x_i)-\mu_{k}^{c})^{T}
\end{equation}
where $N_c$ is the number of training samples of $c-th$ class. We measure the distance between the test sample $x$ and all task Gaussian distributions and select the prefix $p_t$ having the minimum distance. This is done via the Mahalanobis distance as follows: 
\begin{equation}
    c^*=\arg\min_{c\in C} (h(x_i)-\mu_{k}^{c})^{T}A_{k}^{-1}(h(x_i)-\mu_{k}^{c})
\end{equation}
where $c^{*}$ is the selected task. We simply choose the prefix parameters corresponding to the $c^{*}-th$ class. Moreover, we share and accumulate the covariance among all tasks by considering $A=\sum_{k}A_{k}$ to make the ranking calculations more stable \cite{Wang2023RehearsalfreeCL}.

\subsection{Prototype Rectification}
The intra-class bias certainly occurs during the few-shot learning phase due to the data scarcity problem. This issue causes inaccurate prototype calculations because it is performed in such a low-data regime \cite{Liu2019PrototypeRF}. The intra-class bias problem is defined as follows:
\begin{equation}
    B_{intra}=\mathbb{E}_{X^{'}\backsim p_{x^{'}}}[X^{'}]-\mathbb{E}_{X\backsim p_{X}}[X]
\end{equation}
where $p_{X^{'}}$ is the distribution of all data belonging to a certain class, and $p_{X}$ is the distribution of the available labeled data of this class. It is obvious that the deviation of the two distributions is large in the low-data regime. Since the prototype is calculated by averaging all samples, it is safe to interpret the intra-class bias as the difference between the expected prototype drawn by all samples and the actual prototypes computed from those limited available samples. In the realm of \Gls{fscil}, the few-shot learning task is formed in the N-way K-shot setting where $K$ is very small, e.g., 1, 5. This issue causes intra-class bias during the prototype calculation.

To remedy this shortcoming, we train a prediction network $P_{\zeta}(.):\mathcal{Z}\rightarrow\mathcal{\mu}$ inspired by \cite{Xue2020OneShotIC} to perform the prototype completion task. The prediction network $P_{\zeta}(z)$ takes the embedded features $z=f_{\theta}(x)$ and, in turn, maps them to the prototype space. That is, it shares the same embedding of the main network. The target features are the actual prototypes calculated in the base task, while we apply the pseudo-labeling strategy of the testing sets for the few-shot tasks to enrich the sample's representations. On the other hand, input samples are drawn from $\lambda$, the most distant images of the target samples. That is, input samples and target samples are paired to train the prediction network $P_{\zeta}(.)$. In the inference phase, the predicted prototype and the actual prototype are aggregated as follows:
\begin{equation}
    R(\mu)=\frac{1}{2}(P_{\zeta}(\mu)+\mu)
\end{equation}
The refined prototypes $R(\mu)$ are then used to initialize the mean and covariance matrix for that of the Mahalanobis distance calculations as follows:
\begin{equation}
    \mu_c=R(\mu_{c});A=\frac{1}{N_c}\sum_{c}\sum_{y_i=c}(P_{\zeta}(h(x_i)) - \mu_{c})(P_{\zeta}(h(x_i))-\mu_{c})^{T}
\end{equation}
Since the prediction network $P_{\zeta}(.)$ is applied in the continual environments, thus risking the catastrophic forgetting problem, it is structured in multi-model environments where every task is associated with its own prediction network. There does not need any requirements for the oracle during the testing stage because a correct prediction network can be inferred by the task inference module. This facet distinguishes ours from \cite{Xue2020OneShotIC}, where the prediction network is not yet applied in the continual environments.  

\section{Time Complexity Analysis}
The self-supervised learning and supervised learning phases are the most time-consuming stages of the training. Since we backpropagate only through the student model in DINO and there are a constant number of global and local patches, its time complexity is the same as supervised learning. We utilize a \Gls{cct} model for each experiment. The most computationally expensive components of the \Glspl{cct} are the \Gls{mlp} projections in each block and the sequence pooling~\cite{paeedeh2024crossdomain}.\par

We assume that a \Gls{cct} consists of $L$ layers. In each block of the \Gls{cct}, self-attention components perform three matrix calculations. Therefore, for $s$ tokens with the size of $d$ it needs $O(sd^2)$ calculations. Moreover, each \Gls{mlp} with $m$ layer and $n$ neurons in the middle layers requires $O(mn^2)$ calculations. Each convolution layer in the patch embedding component with $c$ layer requires $O(cp^2)$ time, where $p$ is the dimension of the equivalent matrix multiplication. However, these layers are limited to 1-3 layers in practice because of drastic reductions in parameters with each convolution and pooling layer; hence, they are negligible in comparison to the \Gls{mlp} and attention calculations. At last, the sequence pooling has only two matrix operations.\par

In conclusion, the time complexity of a \Gls{cct} model can be summed up to $O(Lsd^2) + O(Lmn^2)$.

\section{Experiments}
This section presents our numerical validation of \Gls{robusta}, encompassing comparisons with prior arts, formal analysis, and ablation studies. 

\subsection{Datasets}
Three benchmark problems, namely CIFAR100, \Gls{mi}, and CUB-200-2011, are utilized here. The CIFAR100 and \acrlong{mi} problems comprise 100 classes, where 60 classes are reserved for the base task while the remainder 40 classes are used for the few-shot tasks under the 5-way-5-shot setting, thus leading to 8 few-shot tasks. For the CUB-200-2011 problem, the base task is built upon 100 classes, while the remaining 100 classes are set for the few-shot tasks under the 10-way-5-shot setting, leading to 10 few-shot tasks. Note that this configuration follows the common setting in the \Gls{fscil} literature \cite{Shi2021OvercomingCF,Kalla2022S3CSS,Chi2022MetaFSCILAM,Tao2020FewShotCL} where the number of base classes for \acrlong{mi}, CIFAR100 and CUB-200-2011 are set to 60, 60 and 100 respectively. In addition, we evaluate the consolidated algorithms with small base tasks where only 20 classes are used to induce the base task of the CIFAR100 problem and the \acrlong{mi} problem, while 50 classes are reserved for the base task of the CUB-200-2011 problem. Numerical results of the 1-shot configuration are also reported here. Table \ref{table: dataset splits} details the dataset splits. 

\subsection{Baseline Algorithms}
\Gls{robusta} is compared against 10 prior arts: iCaRL \cite{Rebuffi2017iCaRLIC}, Rebalance \cite{Hou2019LearningAU}, FSLL \cite{Mazumder2021FewShotLL}, EEIL \cite{Castro2018EndtoEndIL}, F2M \cite{Shi2021OvercomingCF}, MetaFSCIL \cite{Chi2022MetaFSCILAM}, S3C \cite{Kalla2022S3CSS}, FLOWER \cite{Masum2023FewShotCL}, \textcolor{blue}{SSFE-Net \cite{pan2023ssfe} and GKEAL \cite{zhuang2023gkeal}}. Their characteristics are detailed as follows:
\begin{itemize}
    \item iCaRL is a popular memory-based approach using the concept of knowledge distillation. It is chosen for comparison to understand how \Gls{robusta}, having no memory at all, performs compared to a memory-based approach.
    \item EEIL is akin to iCaRL using memory to handle the catastrophic forgetting problem combined with a cross-distilled loss function. It also implements a balanced fine-tuning strategy to address the class's imbalanced situation. Because of the use of memory, EEIL is supposed to be a hard competitor to \Gls{robusta}, having no memory at all. 
    \item Rebalance is another memory-based approach for class-incremental learning. It possesses a balancing strategy to overcome the issue of class imbalance. It is selected for comparison here because it is expected to perform strongly due to the use of memory. 
    \item FSLL is a popular few-shot class-incremental learning strategy adopting the notion of session-trainable parameters leaving other parameters frozen for model updates. The training process is governed by triplet-loss-based metric learning. It follows a similar strategy as \Gls{robusta}, limiting the number of trainable parameters to prevent the over-fitting problem. 
    \item F2M presents a flat learning concept for few-shot class-incremental learning. The flat local region is unveiled in the base learning task and maintained during the few-shot learning task. This method possesses an episodic memory, thus being expected to perform strongly against \Gls{robusta}. 
    \item MetaFSCIL is designed specifically for the few-shot class-incremental learning and applies the bi-level optimization problem to learn quickly from a few samples while addressing the catastrophic forgetting problem. This method can be seen as a \Gls{sota} method in the \Gls{fscil} problem. 
    \item FLOWER is perceived as an extension of F2M where the goal is to establish flat-wide local regions removing the use of memory. This method is selected for comparison because it is equipped with the data augmentation protocol to overcome the data scarcity problem. 
    \item S3C implements the self-supervised learning technique to address the few-shot class-incremental learning problem. As with FLOWER, it possesses a data augmentation protocol that can alleviate the data scarcity problem. 
    \item \textcolor{blue}{SSFE-Net is compared with \Gls{robusta} because it applies the self-supervised learning technique as \Gls{robusta}. Specifically, it applies the knowledge distillation method and the self-supervised learning technique. Note that \Gls{robusta} applies the DINO method as the self-supervised learning technique, whereas the SSFE-Net implements the simCLR method.}
    \item \textcolor{blue}{GKEAL method is deemed as a \Gls{sota} method in the \Gls{fscil} problem where it adopts the kernel technique. It is expected to perform strongly because it utilizes the augmented feature concatenation module.}
\end{itemize}
\textcolor{green}{All algorithms are executed under the same computational resources except for MetaFSCIL, GKEAL, and SSFE-Net, where their hyper-parameters are set with the grid search. MetaFSCIL, GKEAL, and SSFE-Net are not executed under the same computational environments due to the absence of their official source codes. All algorithms are executed five times with different random seeds. The reported results are taken from the average of five runs. Since the source codes of S3C do not facilitate different random seeds and different base classes, it is executed only once for default base classes.}   

\subsection{Implementation Details}
In this subsection, we describe the specifications of all experiments and the architecture of the models we use.\par

We utilized a CCT-14/7x2 for \acrlong{mi} and CCT-21/7x2 for the CIFAR-100 and CUB-200-2011 experiments. Embedding dimensions are 384 for both architectures. We follow \cite{Kalla2022S3CSS} and \cite{Tao2020FewShotCL} implementations to load the datasets. We choose the specified number of classes as base classes for each dataset and divide the remaining classes for the subsequent incremental tasks. The details are displayed in Table \ref{table: dataset splits}.\par 

\begin{table*}[!tb]
    \centering
    \caption{Datasets and tasks specifications}
    \label{table: dataset splits}
    \resizebox{\textwidth}{!}{
        \begin{tabular}{ |>{\color{black}}l|>{\color{black}}p{1.5cm}|>{\color{black}}p{1cm}|>{\color{black}}p{1.5cm}|>{\color{black}}p{1.5cm}|>{\color{black}}p{1.5cm}|>{\color{black}}p{3cm}|>{\color{black}}p{3cm}|>{\color{black}}p{3cm}| }
            \toprule
            Dataset name & Base classes & Ways & Total samples & Samples, train & Samples, test & Train samples, base & Test samples for each task \\
            \bottomrule
            \hline 
            \acrshort{mi} \& CIFAR-100 & 60 & 5 & 60000 & 50000 & 10000 & 30000 & 100 x learned classes \\
            \hline
            \acrshort{mi} \& CIFAR-100 & 20 & 10 & 60000 & 50000 & 10000 & 10000 & 100 x learned classes \\
            \hline
            CUB-200 & 100 & 10 & 11788 & 5994 & 5794 & 3000 & 2864, 3143, 3430, 3728, 4028, 4326, 4614, 4911, 5206, 5494, 5794 \\
            \hline
            CUB-200 & 50 & 15 & 11788 & 5994 & 5794 & 1500 & 1389, 1826, 2272, 2715, 3143, 3578, 4028, 4466, 4911, 5355, 5794 \\
            \hline
        \end{tabular}
    }
\end{table*}

In self-supervised learning, we do linear probing of the frozen teacher model with the Adam optimizer, the learning rate of 1e-3, and batch size of 100 for 200 epochs to prevent overfitting on the base classes with early stopping. Furthermore, the \Gls{sgd} optimizer was used to pre-train the model on ImageNet for the CUB experiments following the TOPIC~\cite{Shi2021OvercomingCF}, S3C~\cite{Kalla2022S3CSS}, and FLOWER~\cite{Masum2023FewShotCL} implementations. In all other cases, we utilize the AdamW optimizer. We use the \Gls{mse} loss function to train the prediction net. In the 1-shot setting, since it is impossible to calculate the Mahalanobis distance for task identification, which is required to be done prior to assigning the pseudo-labels, we utilize the Euclidean distance instead.\par

We do not normalize the samples when loading the dataset because we inserted pixel-wise \Gls{bn} layers in the patch embedding layers of \Glspl{cct}. All samples are resized to 224x224 pixels for all experiments. The weight decay and weight decay end for the cosine scheduler of DINO are 0.04 and 0.4, respectively, for all experiments. Moreover, the teacher temperature and warm-up teacher temperature variables for DINO are set to 0.07 and 0.04, respectively. Prediction networks have 2 layers for \acrlong{mi} and CIFAR-100, but a linear layer was used for the CUB-200 experiments. Table~\ref{table: hyperparameters} shows the remaining notable hyperparameters for all experiments. Other hyperparameters are set to their default values.

\begin{table*}[!t]
    \caption{The list of hyperparameters that are used in different phases of the experiments.}
    \label{table: hyperparameters}
    \centering
    \resizebox{\textwidth}{!}{
        \begin{tabular}{|l|lllp{3cm}|}
            \bottomrule
            \multicolumn{5}{|c|}{Same for all experiments} \\
            \hline
            Phase & Self-supervised learning & Supervised learning & Incremental learning & prediction net (Inc. learning) \\
            \bottomrule
            Learning rate (Classifier) & 0.00025 (Cosine scheduler) & 0.01 & 0.01 & \_ \\
            Learning rate & 0.00025 (Cosine scheduler) & 1e-5 & 0.01 & 1e-3 \\
            Scheduler & Cosine & ReduceLROnPlateau & ReduceLROnPlateau & \_ \\
            Pseudo-labeled samples for task-id stats & \_ & \_ & All tasks & \_ \\
            Pseudo-labeled samples for prediction net & \_ & \_ & \_ & Inc. Tasks \\
            \bottomrule
            \multicolumn{5}{|c|}{\acrlong{mi} (60 base classes)} \\
            \hline
            Phase & Self-supervised learning & Supervised learning & Incremental learning & prediction net (Inc. learning) \\
            \bottomrule
            Batch size & 70 & 230 & 200 & 100 \\
            Patience (ReduceLROnPlateau) & \_ & 10 & 5 & \_ \\
            factor (ReduceLROnPlateau) & \_ & 0.25 & 0.25 & \_ \\
            min\_lr (ReduceLROnPlateau) & \_ & 3e-5 & 0 & \_ \\
            Num. of epochs & 500 (Max) & 1000 (Max) & 4 (base), 15 (inc. tasks) & 300 \\
            Num. of epochs for early stopping & 30 & 30 & \_ & \_ \\
            Weight decay & 0.04 & 3e-6 & 0 & 0 \\
            Prefix seq. length & \_ & \_ & 16 & \_ \\
            Num. of outliers & \_ & \_ & 5 (base), 1 (inc. tasks) & 5 (base), 1 (inc. tasks) \\
            \bottomrule
            \multicolumn{5}{|c|}{\acrlong{mi} (differences in 20 base classes experiments)} \\
            \hline
            Phase & Self-supervised learning & Supervised learning & Incremental learning & prediction net (Inc. learning) \\
            \bottomrule
            Batch size & 70 & 230 & 160 & 100 \\
            Num. of epochs for early stopping & 30 & 80 & \_ & \_ \\
            Num. of epochs & 500 (Max) & 1000 (Max) & 3 (base), 15 (inc. tasks) & 300 \\
            Num. of epochs & 500 (Max) & 1000 (Max) & 4 (base), 15 (inc. tasks) & 100 \\
            \bottomrule
            \multicolumn{5}{|c|}{CIFAR-100 (60 base classes)} \\
            \hline
            Phase & Self-supervised learning & Supervised learning & Incremental learning & prediction net (Inc. learning) \\
            \bottomrule
            Batch size & 120 & 160 & 200 & 100 \\
            Patience (ReduceLROnPlateau) & \_ & 7 & 5 & \_ \\
            factor (ReduceLROnPlateau) & \_ & 0.333 & 0.25 & \_ \\
            min\_lr (ReduceLROnPlateau) & \_ & 3e-5 & 0 & \_ \\
            Num. of epochs & 500 (Max) & 500 (Max) & 2 (base), 15 (inc. tasks) & 500 \\
            Num. of epochs for early stopping & 30 & 30 & \_ & \_ \\
            Weight decay & 0.04 & 1e-5 & 3e-6 & 0 \\
            Prefix seq. length & \_ & \_ & 10 & \_ \\
            Num. of outliers & \_ & \_ & 5 (base), 1 (inc. tasks) & 5 (base), 1 (inc. tasks) \\
            \bottomrule
            \multicolumn{5}{|c|}{CIFAR-100 (differences in 20 base classes experiments)} \\
            \hline
            Phase & Self-supervised learning & Supervised learning & Incremental learning & prediction net (Inc. learning) \\
            \bottomrule
            Batch size & 45 & 160 & 160 & 100 \\
            Num. of epochs & 500 (Max) & 500 (Max) & 3 (base), 15 (inc. tasks) & 100 \\
            Weight decay & 0.04 & 1e-5 & 0 & 0 \\
            Prefix seq. length & \_ & \_ & 16 & \_ \\
            \bottomrule
            \multicolumn{5}{|c|}{CUB-200 (100 base classes)} \\
            \hline
            Phase & Self-supervised learning & Supervised learning & Incremental learning & prediction net (Inc. learning) \\
            \bottomrule
            Batch size & 45 & 150 & 160 & 100 \\
            Patience (ReduceLROnPlateau) & \_ & 10 & 5 & \_ \\
            factor (ReduceLROnPlateau) & \_ & 0.25 & 0.25 & \\
            min\_lr (ReduceLROnPlateau) & \_ & 3e-5 & 0 & \_ \\
            Num. of epochs & 500 (Max) & 1000 (Max) & 5 (base), 10 (inc. tasks) & 500 \\
            Num. of epochs for early stopping & 80 & 80 & \_ & \_ \\
            Weight decay & 0.04 & 1e-5 & 1e-5 & 0 \\
            Prefix seq. length & \_ & \_ & 10 & \_ \\
            Num. of outliers & \_ & \_ & 8 (base), 1 (inc. tasks) & 8 (base), 1 (inc. tasks) \\
            \bottomrule
            \multicolumn{5}{|c|}{CUB-200 (differences in 50 base classes experiments)} \\
            \hline
            Phase & Self-supervised learning & Supervised learning & Incremental learning & prediction net (Inc. learning) \\
            \bottomrule
            Patience (ReduceLROnPlateau) & \_ & 8 & 5 & \_ \\
            Num. of epochs for early stopping & 30 & 30 & \_ & \_ \\
            \hline
        \end{tabular}
    }
\end{table*}

 Finally, in \cite{Kalla2022S3CSS}, they implemented \eqref{eq:reparameterization} with a slight modification \footnote{\href{https://github.com/JAYATEJAK/S3C/blob/main/models/s3c/Network.py}{https://github.com/JAYATEJAK/S3C/blob/main/models/s3c/Network.py}} as follows:
\begin{equation}
\phi=\mu+\mathcal{N}(0,1)\odot Softplus(\sigma - c),
\end{equation}
where $c$ is a hyper-parameter, which was set to 4 in their experiments. We also utilize the same formula in our implementation. On the other side, the hyper-parameters of other algorithms are searched using the grid search technique to ensure fair comparisons. 




\subsection{Preprocessing}
The preprocessing transformations are the same for all datasets. The image size for all \Gls{cct} models is set to 224x224. We perform RandomResizedCrop and RandomHorizontalFlip. However, we do not conduct the Normalize transform like S3C as our model does not require it because of the \Gls{bn} layers. During the testing, we Resize images to (224 x 1.15, 224 x 1.15) and do the CenterCrop. In the base learning phase, we utilize the same transformations in the DINO's implementation, without the Normalize transform. 

\subsection{Numerical Results}
\begin{figure}
    \centering
    \includegraphics[width=\linewidth]{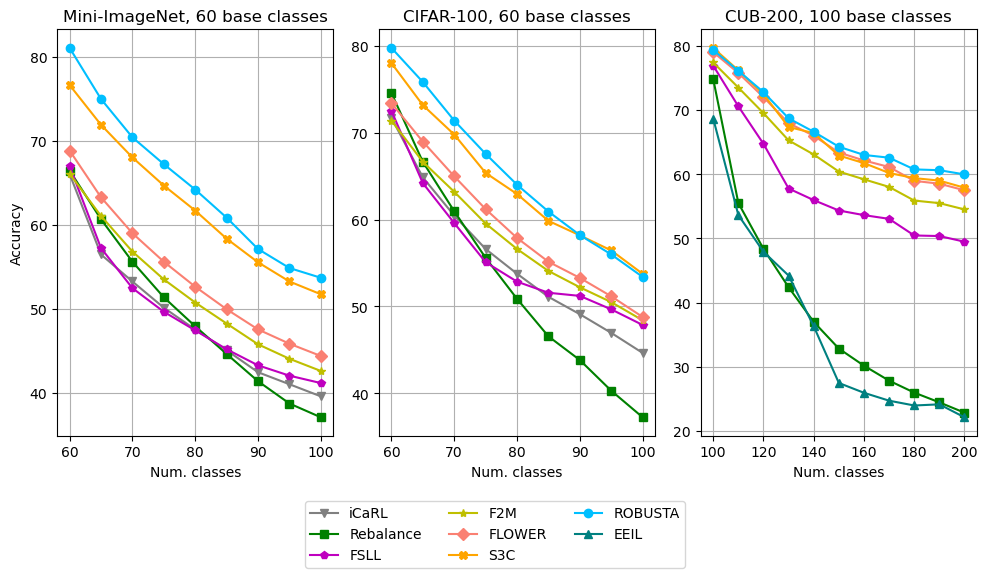}
    \caption{{The trace of the accuracy of the studied methods on the three datasets under moderate base classes (60,60,100)}}
    \label{fig:moderate_number_of_base_classes}
\end{figure}

\begin{figure}
    \centering
    \includegraphics[width=\linewidth]{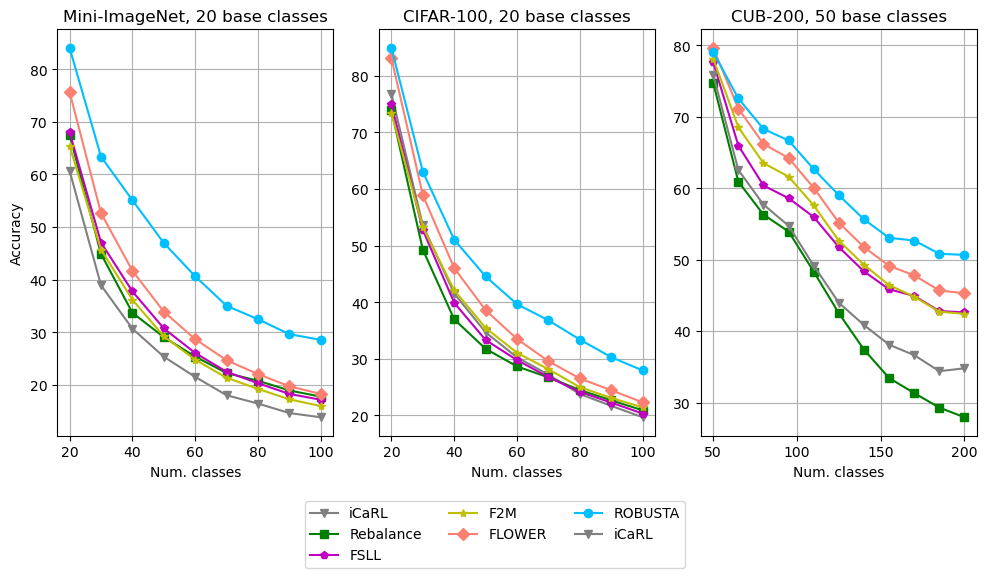}
    \caption{{The trace of the accuracy of the studied methods on the three datasets with small base classes (20,20,50)}}
    \label{fig:fewer_base_classes}
\end{figure}

\begin{figure}
    \centering
    \includegraphics[width=\linewidth]{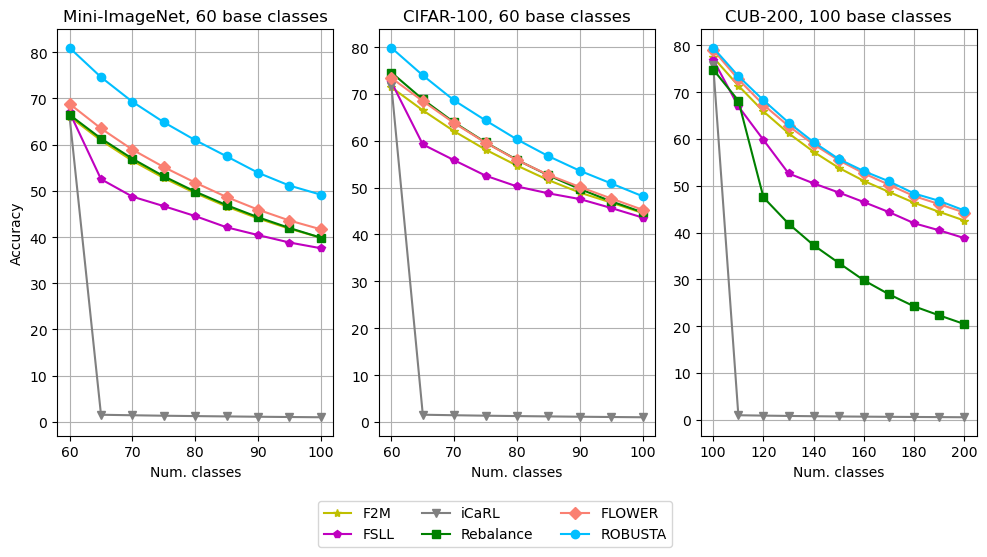}
    \caption{{The trace of the accuracy of the studied methods on the three datasets with moderate base classes and 1-shot setting}}
    \label{fig:1-shot}
\end{figure}

Table \ref{table:fscil_Mini-ImageNet} reports the numerical results of all consolidated algorithms in the \acrlong{mi} problem under $60, 20$ base classes, respectively. The advantage of \Gls{robusta} is clearly seen where it outperforms prior arts with a $2.5\%$ gap to S3C in the second place. The margin becomes wider than that with the $20$ base classes where a $10.96\%$ difference is observed. This finding confirms that \Gls{robusta} is relatively robust against the number of base classes, which happens to be a major problem of the \Gls{fscil} algorithm, i.e., the performances of the \Gls{fscil} algorithms drop significantly in the case of small base classes. The superiority of \Gls{robusta} remains valid for the 1-shot setting as reported in Table \ref{table:fscil_Mini-ImageNet_1_shot} where a $9.38\%$ gap to FLOWER is shown. As with the \acrlong{mi} dataset, \Gls{robusta} outperforms other algorithms in the CIFAR-100 problem with $1.04\%, 5.40\%$ gaps respectively to the second best algorithm for 60 base classes and 20 base classes as reported in Table \ref{table:fscil_CIFAR100}. \Gls{robusta} also beats other algorithms in the 1-shot setting of the CIFAR-100 dataset with $4.39\%$ margin as seen in Table \ref{table:fscil_CIFAR100_1_shot}. Importantly, \Gls{robusta} exceeds S3C with a $1.05\%$ margin with $100$ base classes and FLOWER with $3.18\%$ margin with $50$ base classes in the CUB dataset as shown in Table \ref{table:results_fscil_CUB200}. It delivers a slightly higher accuracy than FLOWER in the second place in the 1-shot setting of the CUB dataset, as exhibited in Table \ref{table:fscil_CUB200_1_shot}. Note that the CUB dataset is deemed challenging because it possesses a small number of samples per class. Fig. \ref{fig:moderate_number_of_base_classes} depicts the trace of classification rates of all consolidated algorithms in the moderate base classes setting, while Fig. \ref{fig:fewer_base_classes} pictorially exhibits the trace of classification rates of all consolidated algorithms in the small base classes setting. Fig. \ref{fig:1-shot} portrays the trace of classification rates of all consolidated algorithms in the 1-shot setting. These figures delineate the advantage of \Gls{robusta}, where it delivers a better accuracy trend than its counterparts. \textcolor{blue}{It is worth noting that the 1-shot case is more challenging than the 5-shot case because only 1 sample per class is available for the training process. This setting usually leads to performance degradation of consolidated algorithms.} \textcolor{red}{Table \ref{table: F1-Score} reports the macros F1-score of \Gls{robusta} across all datasets and settings. It is seen that the difference between the final accuracy and the macros F1-score is small, meaning that the predictions of \Gls{robusta} are not biased toward major classes. Last but not least, Figures 7, 8, and 9, respectively, exhibit the evolution of training losses across all tasks in the moderate base classes, small base classes, and 1-shot setting of the \acrlong{mi} dataset. It is demonstrated that the training losses decrease over time and converge to small points.}

\subsection{Statistical Analysis}
\textcolor{red}{The statistical test is performed for our numerical results to check whether the performance differences between ROBUSTA and baseline algorithms are statistically significant. It is attained using the t-test with a significance level of 0.05. From Table \ref{table:fscil_Mini-ImageNet} to \ref{table:fscil_CUB200_1_shot}, it is seen that ROBUSTA beats other algorithms with statistically significant margins. This, once again, substantiates the superiority of ROBUSTA over other consolidated algorithms.}

\begin{table*}[!tb]
    \centering
    \caption{Average classification accuracy of 5 runs with 5-shot settings on the \acrlong{mi} dataset. The significance level for the t-test is 0.05.}
    \label{table:fscil_Mini-ImageNet}
    \resizebox{\textwidth}{!}{
        \begin{tabular}{|l|lllllllll|l>{\color{blue}}l|>{\color{red}}l>{\color{red}}l|}
            \toprule
            Method & \multicolumn{9}{c}{Accuracy at the end of each task} & & & & \\
            \bottomrule
            Number of base classes & \multicolumn{9}{c}{60 (5-way 5-shot)} &  & & & \\
            \bottomrule
            Task & 0 & 1 & 2 & 3 & 4 & 5 & 6 & 7 & 8 & Avg & Diff. & P-Value & Sig. \\
            \bottomrule
            \ctb iCaRL & \ctb 66.05 & \ctb 56.47 & \ctb 53.26 & \ctb 50.14 & \ctb 47.55 & \ctb 45.08 & \ctb 42.47 & \ctb 41.04 & \ctb 39.60 & \ctb 49.07 & \ctb -15.86 & 8.3e-10 & \checkmark \\
            Rebalance & 66.45 & 60.66 & 55.61 & 51.38 & 47.93 & 44.64 & 41.40 & 38.75 & 37.11 & 49.33 & -15.60 & 5.3e-12 & \checkmark \\
            FSLL & 66.98 & 57.23 & 52.47 & 49.66 & 47.45 & 45.20 & 43.29 & 42.06 & 41.18 & 49.50 & -15.43 & 1.2e-08 & \checkmark \\
            F2M & 66.07 & 61.05 & 56.82 & 53.51 & 50.76 & 48.26 & 45.79 & 44.07 & 42.62 & 52.11 & -12.82 & 2.3e-09 & \checkmark \\
            FLOWER & 68.83 & 63.27 & 59.00 & 55.61 & 52.64 & 49.96 & 47.56 & 45.86 & 44.40 & 54.13 & -10.80 & 1.8e-09 & \checkmark \\
            \ctb SSFE-Net & \ctb 72.06 & \ctb 66.17 & \ctb 62.25 & \ctb 59.74 & \ctb 56.36 & \ctb 53.85 & \ctb 51.96 & \ctb 49.55 & \ctb 47.73 & \ctb 57.74 & \ctb -07.19 & 1.9e-7 & \checkmark \\ 
            MetaFSCIL & 72.04 & 67.94 & 63.77 & 60.29 & 57.58 & 55.16 & 52.9 & 50.79 & 49.19 & 58.85 & -06.08 & 1.7e-06 & \checkmark \\  
            \ctb GKEAL & \ctb 73.59 & \ctb 68.90 & \ctb 65.33 & \ctb 62.29 & \ctb 59.39 & \ctb 56.70 & \ctb 54.20 & \ctb 52.59 & \ctb 51.31 & \ctb 60.48 & \ctb -04.45 & 2.8e-05 & \checkmark \\ 
            S3C & 76.62 & 71.89 & 68.01 & 64.67 & 61.69 & 58.35 & 55.54 & 53.26 & 51.74 & 62.42 & -02.51 & 1.2e-05 & \checkmark \\
            \Gls{robusta} & \textbf{81.04} & \textbf{74.96} & \textbf{70.40} & \textbf{67.21} & \textbf{64.19} & \textbf{60.85} & \textbf{57.12} & \textbf{54.86} & \textbf{53.70} & \textbf{64.93} & \textbf{0} &  & \\
            \bottomrule
            Number of base classes & \multicolumn{9}{c}{20 (10-way 5-shot)} & & & &  \\
            \bottomrule
            Task & 0 & 1 & 2 & 3 & 4 & 5 & 6 & 7 & 8 & Avg & Diff. & P-Value & Sig. \\
            \bottomrule
            F2M & 65.45 & 45.70 & 36.15 & 29.19 & 24.73 & 21.29 & 19.18 & 17.18 & 15.91 & 30.53 & -15.66 & 5.9e-08 & \checkmark \\
            FSLL & 68.05 & 47.05 & 37.74 & 30.70 & 26.00 & 22.43 & 20.26 & 18.22 & 17.11 & 31.95 & -14.24 & 4.7e-08 & \checkmark \\
            iCaRL & 60.70 & 38.91 & 30.65 & 25.30 & 21.46 & 17.97 & 16.38 & 14.58 & 13.77 & 26.64 & -19.55 & 2.5e-07 & \checkmark \\
            Rebalance & 67.55 & 44.89 & 33.67 & 29.04 & 25.32 & 22.19 & 20.73 & 18.88 & 17.70 & 31.11 & -15.08 & 1.2e-06 & \checkmark \\
            FLOWER & 75.70 & 52.66 & 41.67 & 33.88 & 28.64 & 24.63 & 22.00 & 19.67 & 18.21 & 35.23 & -10.96 & 1.6e-08 & \checkmark \\
            \Gls{robusta} & \textbf{84.11} & \textbf{63.40} & \textbf{55.11} & \textbf{46.96} & \textbf{40.61} & \textbf{34.99} & \textbf{32.44} & \textbf{29.59} & \textbf{28.52} & \textbf{46.19} & \textbf{0} & & \\
            \hline
        \end{tabular}
    }
\end{table*}

\begin{table*}[!tb]
    \centering
    \caption{Average classification accuracy of 5 runs with 5-shot settings on the CIFAR-100 dataset. The significance level for the t-test is 0.05.}
    \label{table:fscil_CIFAR100}
    \resizebox{\textwidth}{!}{
        \begin{tabular}{|l|lllllllll|l>{\color{blue}}l|>{\color{red}}l>{\color{red}}l|}
            \toprule
            Method & \multicolumn{9}{c}{Accuracy at the end of each task} & & & & \\
            \bottomrule
            Number of base classes & \multicolumn{9}{c}{60 (5-way 5-shot)} & & & & \\
            \bottomrule
            Task & 0 & 1 & 2 & 3 & 4 & 5 & 6 & 7 & 8 & Avg & Diff. & P-Value & Sig. \\
            \bottomrule
            \ctb iCaRL & \ctb 71.72 & \ctb 64.86 & \ctb 60.46 & \ctb 56.61 & \ctb 53.76 & \ctb 51.10 & \ctb 49.10 & \ctb 46.95 & \ctb 44.64 & \ctb 55.47 & \ctb -09.76 & 1.9e-09 & \checkmark \\
            Rebalance & 74.57 & 66.65 & 60.96 & 55.59 & 50.87 & 46.55 & 43.82 & 40.29 & 37.23 & 52.95 & -12.28 & 3.0e-06 & \checkmark \\
            FSLL & 72.52 & 64.20 & 59.60 & 55.09 & 52.85 & 51.57 & 51.21 & 49.66 & 47.86 & 56.06 & -09.17 & 3.5e-06 & \checkmark \\
            F2M & 71.40 & 66.65 & 63.20 & 59.54 & 56.61 & 54.08 & 52.2 & 50.49 & 48.40 & 58.06 & -07.17 & 1.9e-07 & \checkmark \\
            FLOWER & 73.40 & 68.98 & 64.98 & 61.20 & 57.88 & 55.14 & 53.28 & 51.16 & 48.78 & 59.42 & -05.81 & 1.3e-08 & \checkmark \\
            MetaFSCIL & 74.50 & 70.10 & 66.84 & 62.77 & 59.48 & 56.52 & 54.36 & 52.56 & 49.97 & 60.79 & -04.44 & 7.5e-08 & \checkmark \\  
            \ctb GKEAL & \ctb 74.01 & \ctb 70.45 & \ctb 67.01 & \ctb 63.08 & \ctb 60.01 & \ctb 57.30 & \ctb 55.50 & \ctb 53.39 & \ctb 51.40 & \ctb 61.35 & \ctb -03.88 & 8.6e-06 & \checkmark \\ 
            S3C & 78.02 & 73.18 & 69.79 & 65.37 & 62.94 & 59.85 & 58.18 & 56.44 & 53.75 & 64.17 & -01.06 & 0.01074 & \checkmark \\
            \Gls{robusta} & \textbf{79.77} & \textbf{75.84} & \textbf{71.40} & \textbf{67.56} & \textbf{64.01} & \textbf{60.92} & \textbf{58.21} & {55.98} & {53.39} & \textbf{65.23} & \textbf{0} & & \\ 
            \bottomrule
            Number of base classes & \multicolumn{9}{c}{20 (10-way 5-shot)} & & & & \\
            \bottomrule
            Task & 0 & 1 & 2 & 3 & 4 & 5 & 6 & 7 & 8 & Avg & Diff. & P-Value & Sig. \\
            \bottomrule
            F2M & 73.40 & 53.25 & 42.10 & 35.42 & 31.03 & 28.12 & 24.95 & 23.04 & 21.38 & 36.97 & -08.75 & 4.2e-08 & \checkmark \\
            FSLL & 75.00 & 52.76 & 39.87 & 33.29 & 29.74 & 26.73 & 24.12 & 22.22 & 20.28 & 36.00 & -09.72 & 5.7e-09 & \checkmark \\
            iCaRL & 76.85 & 53.58 & 41.51 & 34.65 & 30.22 & 27.15 & 23.73 & 21.65 & 19.62 & 36.55 & -09.17 & 5.6e-11 & \checkmark \\
            Rebalance & 73.90 & 49.27 & 36.95 & 31.73 & 28.62 & 26.68 & 24.44 & 22.62 & 20.86 & 35.01 & -10.71 & 7.3e-07 & \checkmark \\
            FLOWER & 83.20 & 58.93 & 46.01 & 38.56 & 33.48 & 29.54 & 26.49 & 24.40 & 22.27 & 40.32 & -05.40 & 4.8e-06 & \checkmark \\
            \Gls{robusta} & \textbf{84.96} & \textbf{63.07} & \textbf{51.02} & \textbf{44.53} & \textbf{39.63} & \textbf{36.77} & \textbf{33.37} & \textbf{30.25} & \textbf{27.91} & \textbf{45.72} & \textbf{0} & & \\ 
            \hline
        \end{tabular}
    }
\end{table*}

\begin{table*}[!tb]
    \centering
    \caption{Average classification accuracy of 5 runs with 5-shot settings on the CUB-200 dataset. The significance level for the t-test is 0.05.}
    \label{table:results_fscil_CUB200}
    \resizebox{\textwidth}{!}{
        \begin{tabular}{|p{3cm}|lllllllllll|l>{\color{blue}}l|>{\color{red}}l>{\color{red}}l|}
            \toprule
            Method & \multicolumn{11}{c}{Accuracy at the end of each task} & & & & \\
            \bottomrule
            Number of base classes & \multicolumn{11}{c}{100 (10-way 5-shot)} & &  & &\\
            \bottomrule
            Task & 0 & 1 & 2 & 3 & 4 & 5 & 6 & 7 & 8 & 9 & 10 & Avg & Diff. & P-Value & Sig. \\
            \bottomrule
            Rebalance & 74.83 & 55.51 & 48.36 & 42.38 & 36.98 & 32.82 & 30.15 & 27.82 & 25.96 & 24.44 & 22.83 & 38.37 & -28.43 & 8.4e-07 & \checkmark \\
            FSLL & 76.92 & 70.58 & 64.73 & 57.77 & 55.96 & 54.34 & 53.62 & 53.04 & 50.45 & 50.36 & 49.48 & 57.93 & -08.87 & 2.8e-07 & \checkmark \\
            \ctb iCaRL & \ctb68.68 & \ctb 52.65 & \ctb 48.61 & \ctb 44.16 & \ctb 36.62 & \ctb 29.52 & \ctb 27.83 & \ctb 26.26 & \ctb 24.01 & \ctb 23.89 & \ctb 21.16 & \ctb 36.67 & \ctb 30.13 & 2.0e-07 & \checkmark \\  
            EEIL & 68.68 & 53.63 & 47.91 & 44.2 & 36.3 & 27.46 & 25.93 & 24.7 & 23.95 & 24.13 & 22.11 & 36.27 & -30.53 & 2.4e-07 & \checkmark \\
            \ctb SSFE-Net & \ctb 76.38 & \ctb 72.11 & \ctb 68.82 & \ctb 64.77 & \ctb 63.59 & \ctb 60.56 & \ctb 59.84 & \ctb 58.93 & \ctb 57.33 & \ctb 56.23 & \ctb 54.28 & \ctb 62.99 & \ctb -03.81 & 7.1e-09 & \checkmark \\ 
            MetaFSCIL & 75.90 & 72.41 & 68.78 & 64.78 & 62.96 & 59.99 & 58.30 & 56.85 & 54.78 & 53.82 & 52.64 & 61.92 & -04.88 & 1.6e-07 & \checkmark \\
            \ctb GKEAL & \ctb 78.88 & \ctb 75.62 & \ctb 72.32 & \ctb 68.62 & \ctb \textbf{67.23} & \ctb 64.26 & \ctb 62.98 & \ctb 61.89 & \ctb 60.20 & \ctb 59.21 & \ctb 58.67 & \ctb 66.35 & \ctb -00.45 & 0.01466 & \checkmark \\ 
            F2M & 77.41 & 73.50 & 69.52 & 65.27 & 63.07 & 60.41 & 59.20 & 58.02 & 55.89 & 55.49 & 54.51 & 62.94 & -03.86 & 1.5e-07 & \checkmark \\
            FLOWER & 79.02 & 75.77 & 72.01 & 67.96 & 65.99 & 63.38 & 62.14 & 61.12 & 58.96 & 58.52 & 57.49 & 65.67 & -01.13 & 0.00021 & \checkmark \\
            S3C & \textbf{79.69} & \textbf{76.18} & 72.46 & 67.36 & 66.40 & 62.91 & 61.73 & 60.19 & 59.39 & 59.00 & 57.89 & 65.75 & -01.05 & 0.00143 & \checkmark \\
            \Gls{robusta} & {79.35} & {76.11} & \textbf{72.79} & \textbf{68.71} & {66.61} & \textbf{64.28} & \textbf{63.00} & \textbf{62.57} & \textbf{60.75} & \textbf{60.62} & \textbf{59.97} & \textbf{66.80} & \textbf{0} & & \\
            \bottomrule
            Number of base classes & \multicolumn{11}{c}{50 (15-way 5-shot)} & & & & \\
            \bottomrule
            Task & 0 & 1 & 2 & 3 & 4 & 5 & 6 & 7 & 8 & 9 & 10 & Avg & Diff. & P-Value & Sig. \\
            \bottomrule
            F2M & 78.11 & 68.52 & 63.50 & 61.61 & 57.58 & 52.64 & 49.26 & 46.43 & 44.81 & 42.71 & 42.41 & 55.23 & -05.77 & 2.2e-06 & \checkmark \\
            FSLL & 77.61 & 65.96 & 60.37 & 58.59 & 55.96 & 51.75 & 48.35 & 45.84 & 44.92 & 42.80 & 42.62 & 54.07 & -06.93 & 1.3e-07 & \checkmark \\
            iCaRL & 75.88 & 62.54 & 57.70 & 54.73 & 49.16 & 43.96 & 40.84 & 38.09 & 36.63 & 34.38 & 34.79 & 48.06 & -12.94 & 3.4e-07 & \checkmark \\
            Rebalance & 74.73 & 60.91 & 56.29 & 53.93 & 48.31 & 42.49 & 37.40 & 33.42 & 31.32 & 29.29 & 27.93 & 45.09 & -15.91 & 1.3e-06 & \checkmark \\
            FLOWER & \textbf{79.63} & 71.14 & 66.18 & 64.22 & 60.03 & 55.17 & 51.74 & 49.13 & 47.79 & 45.68 & 45.29 & 57.82 & -03.18 & 7.7e-05 & \checkmark \\
            \Gls{robusta} & {79.05} & \textbf{72.55} & \textbf{68.28} & \textbf{66.67} & \textbf{62.66} & \textbf{59.01} & \textbf{55.63} & \textbf{53.04} & \textbf{52.67} & \textbf{50.81} & \textbf{50.66} & \textbf{61.00} & \textbf{0} & & \\
            \hline
        \end{tabular}
    }
\end{table*}

\begin{table*}[!tb]
    \centering
    \caption{Average classification accuracy of 5 runs with 1-shot settings on the \acrlong{mi} dataset. The significance level for the t-test is 0.05.}
    \label{table:fscil_Mini-ImageNet_1_shot}
    \resizebox{\textwidth}{!}{
        \begin{tabular}{|l|lllllllll|l>{\color{blue}}l|>{\color{red}}l>{\color{red}}l|}
            \toprule
            Method & \multicolumn{9}{c}{Accuracy at the end of each task} & & & & \\
            \bottomrule
            Number of base classes & \multicolumn{9}{c}{60 (5-way 1-shot)} & & & & \\
            \bottomrule
             & 0 & 1 & 2 & 3 & 4 & 5 & 6 & 7 & 8 & Avg & Diff. & P-Value & Sig. \\
            \bottomrule
            F2M & 66.07 & 60.92 & 56.51 & 52.78 & 49.52 & 46.61 & 44.03 & 41.81 & 39.86 & 50.90 & -11.59 & 5.2e-08 & \checkmark \\
            FSLL & 66.98 & 52.53 & 48.75 & 46.69 & 44.54 & 42.11 & 40.43 & 38.82 & 37.58 & 46.49 & -16.00 & 5.2e-07 & \checkmark \\
            iCaRL & 66.05 & 01.54 & 01.43 & 01.33 & 01.25 & 01.18 & 01.11 & 01.05 & 01.00 & 8.44 & -54.05 & 5.6e-06 & \checkmark \\
            Rebalance & 66.45 & 61.33 & 56.95 & 53.15 & 49.83 & 46.90 & 44.29 & 41.96 & 39.86 & 51.19 & -11.30 & 4.0e-08 & \checkmark \\
            FLOWER & 68.83 & 63.50 & 58.92 & 55.14 & 51.78 & 48.66 & 45.93 & 43.58 & 41.67 & 53.11 & -09.38 & 5.9e-08 & \checkmark \\
            \Gls{robusta} & \textbf{80.96} & \textbf{74.61} & \textbf{69.28} & \textbf{64.85} & \textbf{60.97} & \textbf{57.47} & \textbf{53.92} & \textbf{51.14} & \textbf{49.17} & \textbf{62.49} & \textbf{0} & & \\ 
            \hline
        \end{tabular}
    }
\end{table*}

\begin{table*}[!tb]
    \centering
    \caption{Average classification accuracy of 5 runs with 1-shot settings on the CIFAR-100 dataset. The significance level for the t-test is 0.05.}
    \label{table:fscil_CIFAR100_1_shot}
    \resizebox{\textwidth}{!}{
        \begin{tabular}{|l|lllllllll|l>{\color{blue}}l|>{\color{red}}l>{\color{red}}l|}
            \toprule
            Method & \multicolumn{9}{c}{Accuracy at the end of each task} & & & & \\
            \bottomrule
            Number of base classes & \multicolumn{9}{c}{60 (5-way 1-shot)} & & & & \\
            \bottomrule
            & 0 & 1 & 2 & 3 & 4 & 5 & 6 & 7 & 8 & Avg & Diff. & P-Value & Sig. \\
            \bottomrule
            F2M & 71.40 & 66.50 & 62.02 & 58.10 & 54.70 & 51.60 & 48.94 & 46.81 & 44.57 & 56.07 & -05.75 & 2.3e-06 & \checkmark \\
            FSLL & 72.52 & 59.22 & 55.84 & 52.55 & 50.25 & 48.78 & 47.56 & 45.65 & 43.61 & 52.89 & -08.93 & 3.5e-05 & \checkmark \\
            iCaRL & 71.72 & 01.54 & 01.43 & 01.33 & 01.25 & 01.18 & 01.11 & 01.05 & 01.00 & 9.07 & -52.75 & 1.4e-05 & \checkmark \\
            Rebalance & 74.57 & 68.83 & 63.91 & 59.65 & 55.92 & 52.62 & 49.7 & 47.08 & 44.72 & 57.44 & -04.38 & 1.3e-08 & \checkmark \\
            FLOWER & 73.40 & 68.47 & 63.76 & 59.59 & 55.86 & 52.72 & 50.14 & 47.68 & 45.27 & 57.43 & -04.39 & 1.4e-06 & \checkmark \\
            \Gls{robusta} & \textbf{79.84} & \textbf{73.96} & \textbf{68.60} & \textbf{64.34} & \textbf{60.28} & \textbf{56.71} & \textbf{53.60} & \textbf{50.89} & \textbf{48.16} & \textbf{61.82} & \textbf{0} & & \\ 
            \hline
        \end{tabular}
    }
\end{table*}

\begin{table*}[!tb]
    \centering
    \caption{Average classification accuracy of 5 runs with 1-shot settings on the CUB-200 dataset. The significance level for the t-test is 0.05.}
    \label{table:fscil_CUB200_1_shot}
    \resizebox{\textwidth}{!}{
        \begin{tabular}{|p{3cm}|lllllllllll|l>{\color{blue}}l|>{\color{red}}l>{\color{red}}l|}
            \toprule
            Method & \multicolumn{11}{c}{Accuracy at the end of each task} & & & & \\
            \bottomrule
            Task & 0 & 1 & 2 & 3 & 4 & 5 & 6 & 7 & 8 & 9 & 10 & Avg & Diff. & P-Value & Sig. \\
            \bottomrule
            Number of base classes & \multicolumn{11}{c}{100 (10-way 1-shot)}  & & & & \\
            \bottomrule
            F2M & 77.41 & 71.38 & 65.87 & 61.2 & 57.23 & 53.81 & 51.01 & 48.66 & 46.35 & 44.4 & 42.55 & 56.35 & -02.14 & 4.2e-13 & \checkmark \\
            FSLL & 76.92 & 66.97 & 59.78 & 52.62 & 50.49 & 48.54 & 46.49 & 44.35 & 41.97 & 40.46 & 38.79 & 51.58 & -06.91 & 3.1e-07 & \checkmark \\
            iCaRL & 75.73 & 0.95 & 0.87 & 00.80 & 00.74 & 00.69 & 00.65 & 00.61 & 00.58 & 00.55 & 00.52 & 7.52 & -50.97 & 1.4e-06 & \checkmark \\
            Rebalance & 74.83 & 68.18 & 47.5 & 41.87 & 37.27 & 33.49 & 29.77 & 26.78 & 24.21 & 22.28 & 20.43 & 38.78 & -19.71 & 2.5e-06 & \checkmark \\
            FLOWER & 79.02 & 72.73 & 67.29 & 62.74 & 58.82 & 55.43 & 52.63 & 50.18 & 47.77 & 46.05 & 44.26 & 57.90 & -00.59 & 8.9e-07 & \checkmark \\
            \Gls{robusta} & \textbf{79.47} & \textbf{73.42} & \textbf{68.25} & \textbf{63.49} & \textbf{59.29} & \textbf{55.69} & \textbf{53.10} & \textbf{50.92} & \textbf{48.33} & \textbf{46.75} & \textbf{44.69} & \textbf{58.49} & \textbf{0} & & \\ 
            \hline
        \end{tabular}
    }
\end{table*}

\begin{table*}[!tb]
    \centering
    \caption{\ctr F1 scores of \Gls{robusta} for all experiments} 
    \label{table: F1-Score}
    \resizebox{\textwidth}{!}{
        \begin{tabular}{|>{\color{red}}l|>{\color{red}}l|>{\color{red}}l|>{\color{red}}l|>{\color{red}}l|>{\color{red}}l|}
            \toprule
            Dataset & Num. of base classes & Num. of tasks & Shots & Final acc. & F1 Score \\
            \bottomrule
            \acrlong{mi} & 60 & 9 & 5 & 53.40 & 51.82 \\
            \acrlong{mi} & 20 & 9 & 5 & 29.72 & 29.33 \\
            \acrlong{mi} & 60 & 9 & 1 & 49.26 & 41.36 \\
            CIFAR-100 & 60 & 9 & 5 & 53.78 & 50.99 \\
            CIFAR-100 & 20 & 9 & 5 & 28.14 & 25.87 \\
            CIFAR-100 & 60 & 9 & 1 & 48.28 & 38.67 \\
            CUB-200 & 100 & 11 & 5 & 60.20 & 60.66 \\
            CUB-200 & 50 & 11 & 5 & 50.28 & 50.37 \\
            CUB-200 & 100 & 11 & 1 & 43.79 & 39.06 \\
            \bottomrule
        \end{tabular}
    }
    \label{table:f1 score}
\end{table*}




\begin{table*}[!tb]
    \centering
    \caption{Ablation studies on the \acrlong{mi} dataset with 5-shot setting and moderate base classes.}
    \resizebox{\textwidth}{!}{
        \begin{tabular}{|l|lllllllll|l>{\color{blue}}l|}
            \toprule
            Method & \multicolumn{9}{c}{Accuracy at the end of each task} & & \\
            \bottomrule
            Number of base classes & \multicolumn{9}{c}{60 (5-way 5-shot)} &  &  \\
            \bottomrule
            Task & 0 & 1 & 2 & 3 & 4 & 5 & 6 & 7 & 8 & Avg & Diff. \\
            \bottomrule
            \Gls{robusta} & \textbf{81.04} & \textbf{74.96} & \textbf{70.40} & \textbf{67.21} & \textbf{64.19} & \textbf{60.85} & \textbf{57.12} & \textbf{54.86} & \textbf{53.70} & \textbf{64.93} & \textbf{0} \\
            w/o The SSL Phase & {69.79} & {64.33} & {59.36} & {55.74} & {52.34} & {48.96} & {45.42} & {43.26} & {41.45} & {53.41} & {-11.52} \\
            w/o The Prediction Net. & {81.04} & {74.86} & {69.68} & {65.55} & {61.91} & {58.56} & {55.36} & {52.67} & {51.10} & {63.41} & {-01.52} \\
            w/o Stochastic Classifier & {77.58} & {71.61} & {66.51} & {62.08} & {58.24} & {54.80} & {51.76} & {49.03} & {46.59} & {59.80} & {-05.13} \\
            w/o Delta Parameters & {80.06} & {70.19} & {55.74} & {44.09} & {36.98} & {30.78} & {30.25} & {29.51} & {26.62} & {44.91} & {-19.77} \\
            \hline
        \end{tabular}
    }
    \label{table:ablation_studies}
\end{table*}

\begin{figure}
    \centering
    \includegraphics[width=\linewidth]{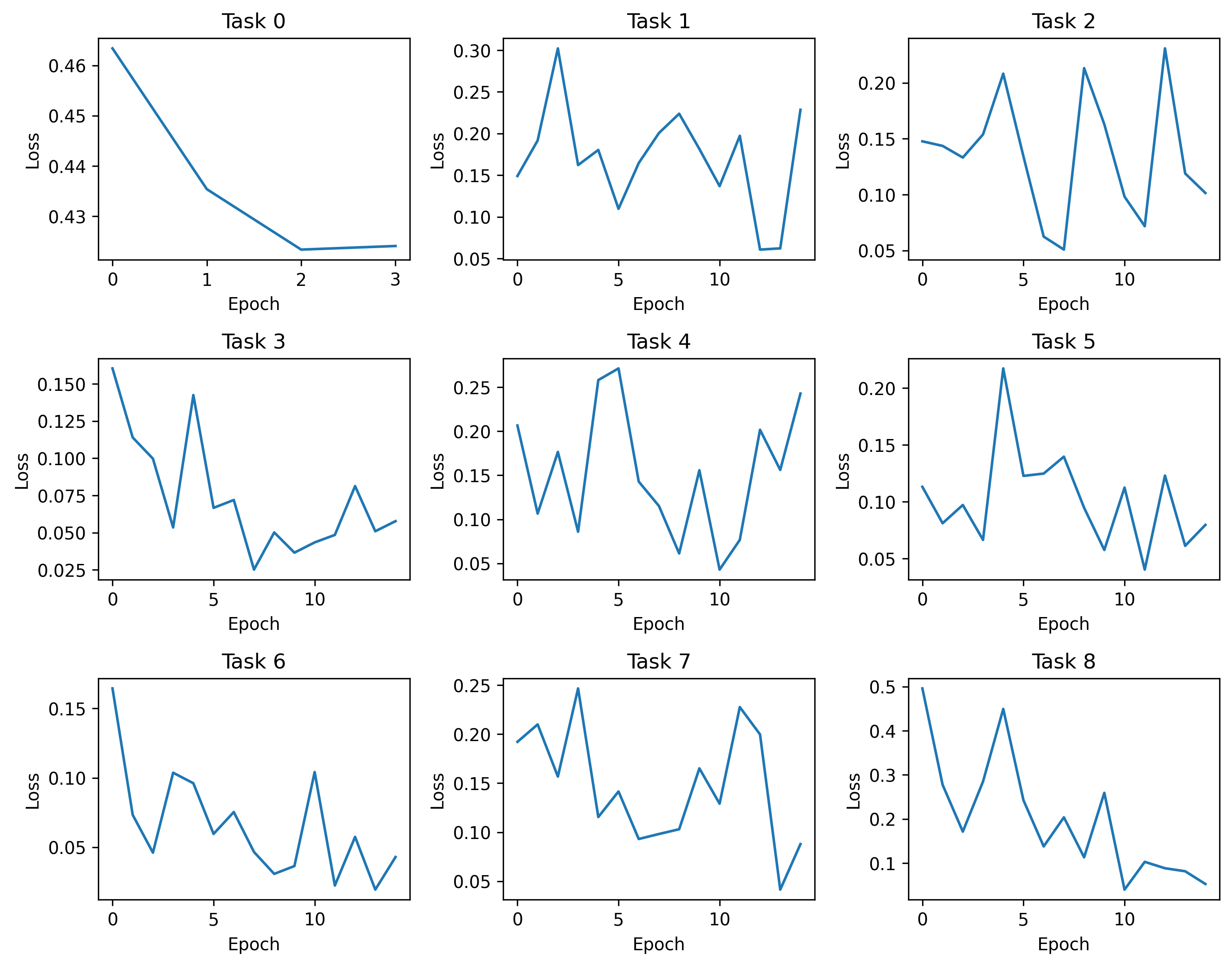}
    \label{fig:loss,mi,60,5}
    \caption{\textcolor{red}{Loss plots for \acrlong{mi} with 60 base classes and 5-shot setting.}}
\end{figure}

\begin{figure}
    \centering
    \includegraphics[width=\linewidth]{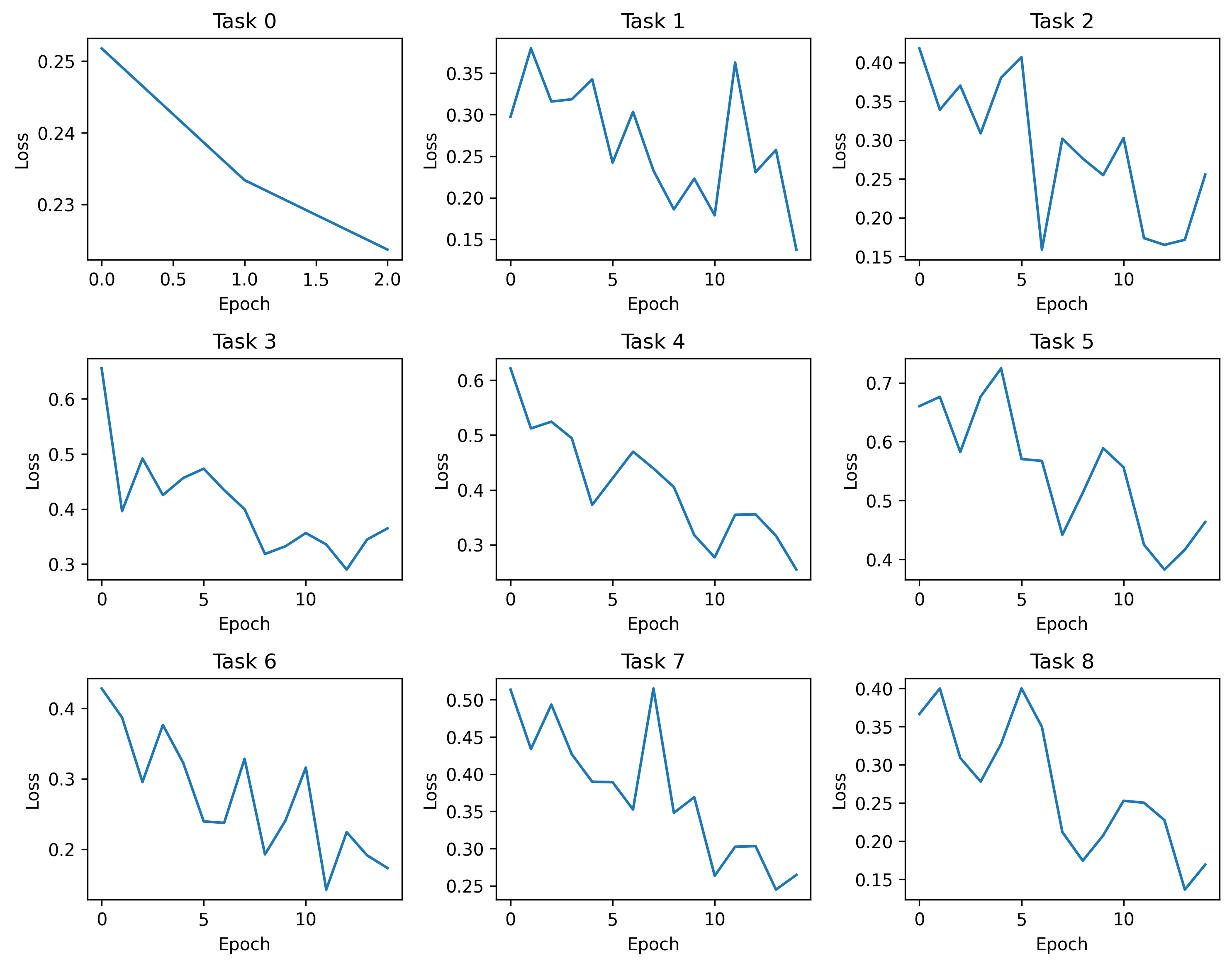}
    \label{fig:loss,mi,20,5}
    \caption{\textcolor{red}{Loss plots for \acrlong{mi} with 20 base classes and 5-shot setting.}}
\end{figure}

\begin{figure}
    \centering
    \includegraphics[width=\linewidth]{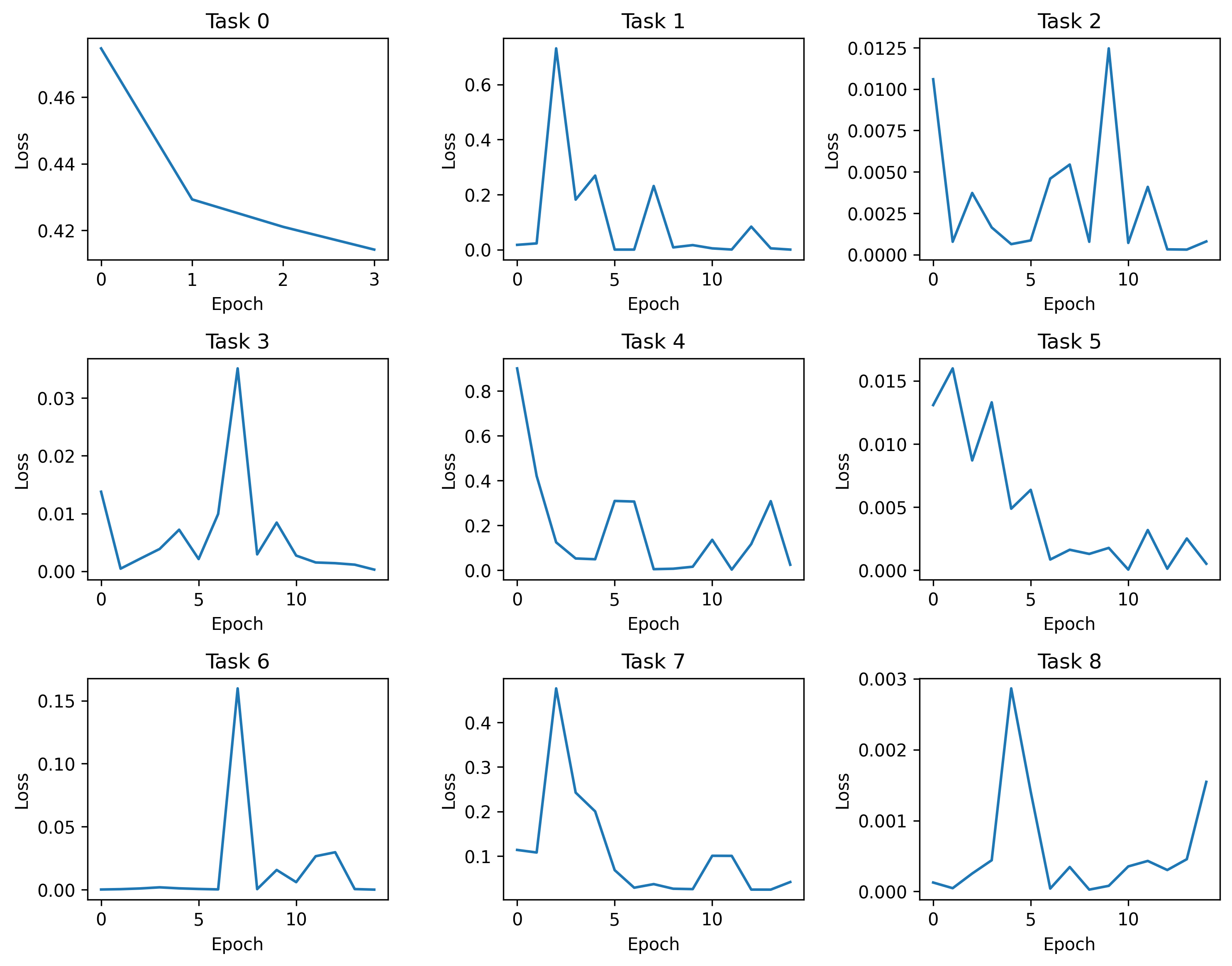}
    \label{fig:loss,mi,60,1}
    \caption{\textcolor{red}{Loss plots for \acrlong{mi} with 60 base classes and 1-shot setting.}}
\end{figure}

\begin{table*}[!tb]
    \centering
    \caption{Average forgetting of \Gls{robusta} and S3C on \acrlong{mi}, CIFAR-100, and CUB-200 datasets for 5-shot setting.}
    \label{table:average forgetting}
    \resizebox{0.8 \textwidth}{!}{
        \begin{tabular}{ |>{\color{black}}l|>{\color{black}}l|>{\color{black}}l|>{\color{black}}l|>{\color{black}}l|}
            \toprule
            Dataset name & Method & Base classes & Average forgetting and standard deviation \\
            \bottomrule
            \hline
            \acrlong{mi} & S3C     & 60 & 3.52 (1 run) \\
            \acrlong{mi} & \Gls{robusta} & 60 & 3.54, Std: 1.61 (5 runs) \\
            \hline
            CIFAR-100 & S3C     & 60 & 5.18 (1 run) \\
            CIFAR-100 & \Gls{robusta} & 60 & 1.39, Std: 0.42 (5 runs) \\
            \hline
            CUB-200 & S3C     & 200 & 5.64 (1 run) \\
            CUB-200 & \Gls{robusta} & 200 & 2.44, Std: 0.61 (5 runs) \\
            \hline
        \end{tabular}
    }
\end{table*}

\begin{table*}[!tb]
    \centering
    \caption{Comparison of the number of parameters between \Gls{robusta} and S3C on \acrlong{mi}, CIFAR-100, and CUB-200 datasets for 5-shot setting.}
    \label{table:parameters}
    \resizebox{0.7 \textwidth}{!}{
        \begin{tabular}{ |>{\color{black}}l|>{\color{black}}l|>{\color{black}}l|>{\color{black}}l|}
            \toprule
            Dataset name & Method & Total parameters & Learnable parameters \\
            \bottomrule
            \hline
            \acrlong{mi} & S3C      & 12.64 M & 409.6 K \\
            \hline
            \acrlong{mi} & \Gls{robusta}  & 29.27 M & 4.29 M   \\
            \hline
            CIFAR-100 & S3C         & 339.6 K & 51.2 K \\   
            \hline
            CIFAR-100 & \Gls{robusta}     & 39.5 M & 4.12 M \\
            \hline
            CUB-200 & S3C           & 13.05 M & 819.2 K \\  
            \hline
            CUB-200 & \Gls{robusta}       & 38.62 M & 3.24 M \\
            \hline
        \end{tabular}
    }
\end{table*}

\begin{table*}[!tb]
    \centering
    \caption{Execution time of the elapsed time between \Gls{robusta} and S3C on \acrlong{mi}, CIFAR-100, and CUB-200 datasets for incremental tasks for 5-shot setting.}
    \label{table:execution_time_incremental_tasks}
    \resizebox{0.6 \textwidth}{!}{
        \begin{tabular}{ |>{\color{black}}l|>{\color{black}}l|>{\color{black}}p{5cm}|>{\color{black}}p{5cm}|}
            \toprule
            Dataset name & Method & Execution time \\
            \bottomrule
            \hline
            \acrlong{mi} & S3C      & 08 min, 19 sec \\
            \hline
            \acrlong{mi} & \Gls{robusta}  & 36 min, 01 sec \\
            \hline
            CIFAR-100 & S3C         & 05 min, 32 sec \\
            \hline
            CIFAR-100 & \Gls{robusta}     & 42 min, 54 sec \\
            \hline
            CUB-200 & S3C           & 18 min, 20 sec \\
            \hline
            CUB-200 & \Gls{robusta}       & 30 min, 07 sec \\
            \hline
        \end{tabular}
    }
\end{table*}

\begin{table*}[!tb]
    \centering
    \caption{Sensitivity analysis on the impact of the optimizer and learning rate on \acrlong{mi} with 60 base classes and 5-shot setting.}
    \label{table:sensitivity_analysis}
    \resizebox{0.7 \textwidth}{!}{
        \begin{tabular}{ |>{\color{black}}l|>{\color{black}}l|>{\color{black}}l|>{\color{black}}l|>{\color{black}}l|>{\color{black}}l|}
            \toprule
            Optimizer & lr (model) & lr (prediction net) & Task 1 & Task 9 & Avg. \\
            \bottomrule
            \hline
            AdamW & 1e-3 & 1e-4 & 80.57 & 54.63 & 65.14 \\
            \hline
            AdamW & 1e-3 & 1e-3 & 80.57 & 52.51 & 64.35 \\
            \hline
            AdamW & 1e-3 & 1e-2 & 80.57 & 50.81 & 63.63 \\
            \hline
            AdamW & 1e-2 & 1e-4 & 80.93 & 54.66 & 65.24 \\
            \hline
            \textbf{AdamW (5 runs)} & 1e-2 & 1e-3 & 81.04 & 53.70 & 64.93 \\
            \hline
            AdamW & 1e-2 & 1e-2 & 80.93 & 52.30 & 64.25 \\
            \hline
            AdamW & 1e-1 & 1e-4 & 80.42 & 27.47 & 52.27 \\
            \hline
            AdamW & 1e-1 & 1e-3 & 80.42 & 25.62 & 50.96 \\
            \hline
            AdamW & 1e-1 & 1e-2 & 80.42 & 28.85 & 52.79 \\
            \hline
            Adam & 1e-4 & 1e-4 & 80.03 & 54.08 & 64.49 \\
            \hline
            Adam & 1e-4 & 1e-3 & 80.03 & 52.42 & 64.01 \\
            \hline
            Adam & 1e-4 & 1e-2 & 80.03 & 51.10 & 63.38 \\
            \hline
            Adam & 1e-3 & 1e-4 & 80.57 & 54.63 & 65.14 \\
            \hline
            Adam & 1e-3 & 1e-3 & 80.57 & 52.51 & 64.35 \\
            \hline
            Adam & 1e-3 & 1e-2 & 80.57 & 50.81 & 63.63 \\
            \hline
            Adam & 1e-2 & 1e-4 & 80.93 & 54.66 & 65.24 \\
            \hline
            Adam & 1e-2 & 1e-3 & 80.93 & 53.40 & 64.77 \\
            \hline
            Adam & 1e-2 & 1e-2 & 80.93 & 52.30 & 64.25 \\
            \hline
            SGD & 1e-4 & 1e-4 & 80.02 & 56.56 & 64.95 \\
            \hline
            SGD & 1e-4 & 1e-3 & 80.02 & 55.92 & 65.26 \\
            \hline
            SGD & 1e-4 & 1e-2 & 80.02 & 56.56 & 64.95 \\
            \hline
            SGD & 1e-3 & 1e-4 & 80.02 & 56.51 & 64.95 \\
            \hline
            SGD & 1e-3 & 1e-3 & 80.02 & 56.13 & 65.31 \\
            \hline
            SGD & 1e-3 & 1e-2 & 80.02 & 54.37 & 64.87 \\
            \hline
            SGD & 1e-2 & 1e-4 & 80.05 & 56.46 & 64.93 \\
            \hline
            SGD & 1e-2 & 1e-3 & 80.05 & 55.92 & 65.27 \\
            \hline
            SGD & 1e-2 & 1e-2 & 80.05 & 54.32 & 64.86 \\
            \hline
        \end{tabular}
    }
\end{table*}

\subsection{Ablation Study}
This section discusses the ablation study of \Gls{robusta} done with the \acrlong{mi} dataset in the 5-way 5-shot setting where numerical results are presented in Table \ref{table:ablation_studies}. It is perceived that the absence of the self-supervised learning phase via DINO results in the performance degradation of \Gls{robusta} by over $10\%$. This finding confirms the advantage of DINO to induce meaningful representations enabling the base model to generalize beyond the base classes. The same finding is found with the removal of the prediction network for the prototype rectifications. It causes performance drops of \Gls{robusta} by over $1\%$. This reduction is attributed to inaccurate prototype estimations of \Gls{robusta} due to very low samples, i.e., 5 samples per class. The importance of stochastic classifier is confirmed where its absence deteriorates the performance of \Gls{robusta} by over $5\%$. The stochastic classifier underpins the application of infinite classifiers where the classifier weights are sampled from a distribution. Last but not least, the delta parameters play vital roles in combating the \Gls{cf} problem. Its absence reduces the accuracy of \Gls{robusta}. This finding substantiates our claims that each learning component of \Gls{robusta} contributes positively.  Several facets are observed in the ablation study. 
\begin{itemize}
    \item The self-supervised learning phase plays a vital role in boosting the performance of \Gls{robusta} in the base task. Its absence significantly compromises the performance of \Gls{robusta} in the base task. 
    \item The prediction network actually works to address the intra-class bias problem due to the problem of data scarcity in the few-shot learning tasks. The few-shot learning task accuracy of \Gls{robusta} drops with the absence of the prediction network. 
    \item The stochastic classifier is important to \Gls{robusta}, where its absence brings down the performance of \Gls{robusta} for both the base task and the few-shot learning tasks. This observation is related to the issues of data scarcity and catastrophic forgetting, which can be alleviated with the use of a stochastic classifier. 
    \item The delta parameters contribute to the solution of the catastrophic forgetting where its absence degenerates the accuracy by about $19.77\%$. 
\end{itemize}

\subsection{Analysis of Forgetting}
The issue of \Gls{cf} is analyzed in this section using the average forgetting metric \cite{Chaudhry2019EfficientLL}. \Gls{robusta} is compared with S3C \cite{Kalla2022S3CSS} and Table \ref{table:average forgetting} reports our numerical results. Note that the forgetting analysis is important because a high accuracy is not always equivalent to a low forgetting. It is seen from Table \ref{table:average forgetting} that \Gls{robusta} delivers lower forgetting than S3C in the CIFAR-100 and CUB-200 datasets, while it is comparable to S3C in the \acrlong{mi} dataset. This fact implies that the drops in average accuracy in each session are less in \Gls{robusta} than that in S3C. On the other hand, the average forgetting is relatively small in \Gls{robusta}, meaning that the drops in average accuracy per session are also small. 

\subsection{Analysis of Network Parameters}
Table \ref{table:parameters} tabulates the number of network parameters of \Gls{robusta} and S3C. It is seen that the number of parameters of \Gls{robusta} is not equivalent to the number of learnable parameters because of the use of delta parameters. That is, only delta parameters and prediction networks are learnable during the few-shot learning phase leaving other parameters frozen. This strategy is to avoid the overfitting problem and the catastrophic forgetting problem. Compared to S3C, \Gls{robusta} imposes higher numbers of network parameters than S3C because S3C is constructed under the convolutional backbone, i.e., ResNet. 

\subsection{Analysis of Execution Times}
Table \ref{table:execution_time_incremental_tasks} details the execution times of \Gls{robusta} and S3C across the three datasets under the 1-shot setting. It is perceived that the execution times of \Gls{robusta} are higher than S3C. This fact is understandable because \Gls{robusta} incorporates additional learning components compared to S3C. In addition, \Gls{robusta} makes use of more complex backbone networks than S3C, calling for extra network parameters affecting the execution times. \Gls{robusta}, however, maintains superiority over S3C in terms of accuracy, where it beats S3C in all cases.

\subsection{Sensitivity Analysis}
Table \ref{table:sensitivity_analysis} reports the numerical results of our sensitivity analysis where the learning rates of the model and the prediction network are varied while attempting different optimization strategies. It is seen from Table \ref{table:sensitivity_analysis} that \Gls{robusta} is robust against variations in learning rates and optimization strategies. That is, performance differences are marginal with different learning rates and optimization strategies. This finding is crucial to show the advantage of \Gls{robusta} where our numerical results are not found by chance.

\subsection{Discussion}
Six aspects are observed from our experiments:
\begin{enumerate}
    \item The \Gls{fscil} is a highly challenging problem and features three major issues: over-fitting, catastrophic forgetting, and intra-class bias. To succeed in the \Gls{fscil}, the three problems have to be tackled simultaneously. 
    \item \Gls{robusta} is capable of outperforming other algorithms in all cases where high margins are found in the small base task setting and in the 1-shot setting. This observation is underpinned by the fact that the issues of over-fitting and intra-class bias are severe in such a setting. 
    \item \Gls{robusta}, even without any data augmentation protocol, beats other algorithms. This finding is mainly attributed to the transformer backbone of \Gls{robusta} producing high base task accuracy. 
    \item The key difference between \Gls{robusta} and other algorithms lies in the intra-class bias problem excluded in other algorithms resulting in inaccurate prototype calculation. \Gls{robusta} applies the prototype rectification strategy, which alleviates the intra-class bias problem.
    \item Execution times of \Gls{robusta} are higher than S3C. This issue is understandable because \Gls{robusta} incorporates more learning components than S3C imposing additional execution time. On the other hand, notwithstanding that \Gls{robusta} utilizes a complex backbone network, it does not suffer from the over-fitting problem because the number of learnable parameters is relatively small, i.e., only the delta parameters are learned, leaving the backbone network frozen.
    \item \textcolor{red}{Another advantage of \Gls{robusta} lies in the absence of any data augmentation protocol. This feature is desired in the limited computational resources where the sample augmentation operation imposes additional computational resources. Besides, the data augmentation procedure is linked to the out of distribution problem \cite{Masum2023AssessorGuidedLF} suppressing the generalization potential.}
\end{enumerate}

\subsection{Limitations}
Although \Gls{robusta} performs satisfactorily compared to the other algorithms for the \Gls{fscil} problems, there is still some room for improvement:
\begin{itemize}
    \item \Gls{robusta} utilizes a complex backbone network and a prediction network incurring a high number of parameters. This hinders its deployments in limited computational resources environments. This issue affects the execution times of \Gls{robusta}, imposing higher execution times than S3C, as shown in Table \ref{table:execution_time_incremental_tasks}.  
    \item \Gls{robusta} utilizes the Mahalanobis distance for task inference, which might be inaccurate sometimes. This may lead to a drop in accuracy performance because wrong delta parameters are assigned. 
    \item \Gls{robusta} relies on the combination of the supervised learning phase and the self-supervised learning phase in the base learning task. This strategy imposes considerable computational complexity because DINO entails significant amounts of resources and time to be executed. 
\end{itemize}

\section{Conclusion}
This paper proposes a transformer solution for \Gls{fscil} problems, \Gls{robusta}. \Gls{robusta} is developed from the construct of \Gls{cct} incorporating the batch norm layer to stabilize the training process and the stochastic classifier to overcome the over-fitting problem as a result of the data scarcity problem. The catastrophic forgetting problem is coped with the notion of delta parameters with a non-parametric approach for task inference, while the problem of intra-class bias is handled with the deployment of a prediction network for prototype refinements. Our numerical results demonstrate the advantage of \Gls{robusta}, where it outperforms other algorithms in all cases with noticeable margins. Significant gaps are found in the small base task and 1-shot settings where the problems of over-fitting and intra-class bias are severe. In addition, the ablation study substantiates the advantage of each learning module of \Gls{robusta} while the analysis of forgetting reveals that our algorithm incurs less forgetting than the prior art. The limitation of \Gls{robusta} exists in the issue of complexity, where it imposes higher parameters and execution times than those of prior arts. Although \Gls{robusta} is relatively successful in handling the \Gls{fscil} problem, it still excludes the problem of domain shifts where all tasks are derived from the same domain. Our future work is devoted to studying the topic of cross-domain few-shot class-incremental learning, where each task is drawn from different domains. This problem requires a model to align the distributions of each domain in addition to the solutions of the three major problems of the \Gls{fscil}, overfitting, intra-class bias, \Gls{cf}.

\bibliographystyle{elsarticle-num} 
\bibliography{references}

\begin{thebibliography}{10}
\expandafter\ifx\csname url\endcsname\relax
  \def\url#1{\texttt{#1}}\fi
\expandafter\ifx\csname urlprefix\endcsname\relax\def\urlprefix{URL }\fi
\expandafter\ifx\csname href\endcsname\relax
  \def\href#1#2{#2} \def\path#1{#1}\fi

\bibitem{Tao2020FewShotCL}
X.~Tao, X.~Hong, X.~Chang, S.~Dong, X.~Wei, Y.~Gong, Few-shot class-incremental learning, 2020 IEEE/CVF Conference on Computer Vision and Pattern Recognition (CVPR) (2020) 12180--12189.

\bibitem{Parisi2019ContinualLL}
G.~I. Parisi, R.~Kemker, J.~L. Part, C.~Kanan, S.~Wermter, Continual lifelong learning with neural networks: A review, Neural networks : the official journal of the International Neural Network Society 113 (2019) 54--71.

\bibitem{Mazumder2021FewShotLL}
P.~Mazumder, P.~Singh, P.~Rai, Few-shot lifelong learning, in: AAAI, 2021.

\bibitem{Chen2021IncrementalFL}
K.~Chen, C.-G. Lee, Incremental few-shot learning via vector quantization in deep embedded space, in: ICLR, 2021.

\bibitem{Zhang2021FewShotIL}
C.~Zhang, N.~Song, G.~Lin, Y.~Zheng, P.~Pan, Y.~Xu, Few-shot incremental learning with continually evolved classifiers, 2021 IEEE/CVF Conference on Computer Vision and Pattern Recognition (CVPR) (2021) 12450--12459.

\bibitem{Shi2021OvercomingCF}
G.~Shi, J.~Chen, W.~Zhang, L.-M. Zhan, X.-M. Wu, Overcoming catastrophic forgetting in incremental few-shot learning by finding flat minima, in: NeurIPS, 2021.

\bibitem{Kalla2022S3CSS}
J.~Kalla, S.~Biswas, S3c: Self-supervised stochastic classifiers for few-shot class-incremental learning, in: European Conference on Computer Vision, 2022.

\bibitem{Lu2020StochasticCF}
Z.~Lu, Y.~Yang, X.~Zhu, C.~Liu, Y.-Z. Song, T.~Xiang, Stochastic classifiers for unsupervised domain adaptation, 2020 IEEE/CVF Conference on Computer Vision and Pattern Recognition (CVPR) (2020) 9108--9117.

\bibitem{Hassani2021EscapingTB}
A.~Hassani, S.~Walton, N.~Shah, A.~Abuduweili, J.~Li, H.~Shi, Escaping the big data paradigm with compact transformers, ArXiv abs/2104.05704 (2021).

\bibitem{Yao2021LeveragingBN}
Z.~Yao, Y.~Cao, Y.~Lin, Z.~Liu, Z.~Zhang, H.~Hu, Leveraging batch normalization for vision transformers, 2021 IEEE/CVF International Conference on Computer Vision Workshops (ICCVW) (2021) 413--422.

\bibitem{Wang2021LearningTP}
Z.~Wang, Z.~Zhang, C.-Y. Lee, H.~Zhang, R.~Sun, X.~Ren, G.~Su, V.~Perot, J.~G. Dy, T.~Pfister, \href{https://api.semanticscholar.org/CorpusID:245218925}{Learning to prompt for continual learning}, 2022 IEEE/CVF Conference on Computer Vision and Pattern Recognition (CVPR) (2021) 139--149.
\newline\urlprefix\url{https://api.semanticscholar.org/CorpusID:245218925}

\bibitem{Wang2022DualPromptCP}
Z.~Wang, Z.~Zhang, S.~Ebrahimi, R.~Sun, H.~Zhang, C.-Y. Lee, X.~Ren, G.~Su, V.~Perot, J.~G. Dy, T.~Pfister, \href{https://api.semanticscholar.org/CorpusID:248085201}{Dualprompt: Complementary prompting for rehearsal-free continual learning}, ArXiv abs/2204.04799 (2022).
\newline\urlprefix\url{https://api.semanticscholar.org/CorpusID:248085201}

\bibitem{Wang2023RehearsalfreeCL}
Z.~Wang, Y.~Liu, T.~Ji, X.~Wang, Y.~Wu, C.~Jiang, Y.~Chao, Z.~Han, L.~Wang, X.~Shao, W.~Zeng, \href{https://api.semanticscholar.org/CorpusID:259370817}{Rehearsal-free continual language learning via efficient parameter isolation}, in: Annual Meeting of the Association for Computational Linguistics, 2023.
\newline\urlprefix\url{https://api.semanticscholar.org/CorpusID:259370817}

\bibitem{Liu2019PrototypeRF}
J.~Liu, L.~Song, Y.~Qin, \href{https://api.semanticscholar.org/CorpusID:208267646}{Prototype rectification for few-shot learning}, in: European Conference on Computer Vision, 2019.
\newline\urlprefix\url{https://api.semanticscholar.org/CorpusID:208267646}

\bibitem{Xue2020OneShotIC}
W.~Xue, W.~Wang, \href{https://api.semanticscholar.org/CorpusID:213656799}{One-shot image classification by learning to restore prototypes}, in: AAAI Conference on Artificial Intelligence, 2020.
\newline\urlprefix\url{https://api.semanticscholar.org/CorpusID:213656799}

\bibitem{Caron2021EmergingPI}
M.~Caron, H.~Touvron, I.~Misra, H.~J'egou, J.~Mairal, P.~Bojanowski, A.~Joulin, Emerging properties in self-supervised vision transformers, 2021 IEEE/CVF International Conference on Computer Vision (ICCV) (2021) 9630--9640.

\bibitem{Doersch2020CrossTransformersSF}
C.~Doersch, A.~Gupta, A.~Zisserman, Crosstransformers: spatially-aware few-shot transfer, ArXiv abs/2007.11498 (2020).

\bibitem{Kirkpatrick2017OvercomingCF}
J.~Kirkpatrick, R.~Pascanu, N.~C. Rabinowitz, J.~Veness, G.~Desjardins, A.~A. Rusu, K.~Milan, J.~Quan, T.~Ramalho, A.~Grabska-Barwinska, D.~Hassabis, C.~Clopath, D.~Kumaran, R.~Hadsell, Overcoming catastrophic forgetting in neural networks, Proceedings of the National Academy of Sciences 114 (2017) 3521 -- 3526.

\bibitem{Zenke2017ContinualLT}
F.~Zenke, B.~Poole, S.~Ganguli, Continual learning through synaptic intelligence, Proceedings of machine learning research 70 (2017) 3987--3995.

\bibitem{Aljundi2018MemoryAS}
R.~Aljundi, F.~Babiloni, M.~Elhoseiny, M.~Rohrbach, T.~Tuytelaars, Memory aware synapses: Learning what (not) to forget, in: ECCV, 2018.

\bibitem{Cha2021CPRCR}
S.~Cha, H.~Hsu, F.~du~Pin~Calmon, T.~Moon, Cpr: Classifier-projection regularization for continual learning, ArXiv abs/2006.07326 (2021).

\bibitem{Paik2020OvercomingCF}
I.~Paik, S.~Oh, T.~Kwak, I.~Kim, Overcoming catastrophic forgetting by neuron-level plasticity control, ArXiv abs/1907.13322 (2020).

\bibitem{Li2016LearningWF}
Z.~Li, D.~Hoiem, Learning without forgetting, IEEE Transactions on Pattern Analysis and Machine Intelligence 40 (2016) 2935--2947.

\bibitem{Mao2021ContinualLV}
F.~Mao, W.~Weng, M.~Pratama, E.~Yapp, Continual learning via inter-task synaptic mapping, ArXiv abs/2106.13954 (2021).

\bibitem{Rusu2016ProgressiveNN}
A.~A. Rusu, N.~C. Rabinowitz, G.~Desjardins, H.~Soyer, J.~Kirkpatrick, K.~Kavukcuoglu, R.~Pascanu, R.~Hadsell, Progressive neural networks, ArXiv abs/1606.04671 (2016).

\bibitem{Yoon2018LifelongLW}
J.~Yoon, E.~Yang, J.~Lee, S.~J. Hwang, Lifelong learning with dynamically expandable networks, ArXiv abs/1708.01547 (2018).

\bibitem{Li2019LearnTG}
X.~lai Li, Y.~Zhou, T.~Wu, R.~Socher, C.~Xiong, Learn to grow: A continual structure learning framework for overcoming catastrophic forgetting, in: ICML, 2019.

\bibitem{Xu2021AdaptivePC}
J.~Xu, J.~Ma, X.~Gao, Z.~Zhu, Adaptive progressive continual learning., IEEE transactions on pattern analysis and machine intelligence PP (2021).

\bibitem{Ashfahani2022UnsupervisedCL}
A.~Ashfahani, M.~Pratama, Unsupervised continual learning in streaming environments, IEEE transactions on neural networks and learning systems PP (2022).

\bibitem{Pratama2021UnsupervisedCL}
M.~Pratama, A.~Ashfahani, E.~Lughofer, Unsupervised continual learning via self-adaptive deep clustering approach, ArXiv abs/2106.14563 (2021).

\bibitem{Zoph2017NeuralAS}
B.~Zoph, Q.~V. Le, Neural architecture search with reinforcement learning, ArXiv abs/1611.01578 (2017).

\bibitem{Rakaraddi2022ReinforcedCL}
A.~Rakaraddi, S.-K. Lam, M.~Pratama, M.~V. de~Carvalho, Reinforced continual learning for graphs, Proceedings of the 31st ACM International Conference on Information \& Knowledge Management (2022).

\bibitem{Rebuffi2017iCaRLIC}
S.-A. Rebuffi, A.~Kolesnikov, G.~Sperl, C.~H. Lampert, icarl: Incremental classifier and representation learning, 2017 IEEE Conference on Computer Vision and Pattern Recognition (CVPR) (2017) 5533--5542.

\bibitem{Castro2018EndtoEndIL}
F.~M. Castro, M.~J. Mar{\'i}n-Jim{\'e}nez, N.~G. Mata, C.~Schmid, A.~Karteek, End-to-end incremental learning, ArXiv abs/1807.09536 (2018).

\bibitem{Hou2019LearningAU}
S.~Hou, X.~Pan, C.~C. Loy, Z.~Wang, D.~Lin, Learning a unified classifier incrementally via rebalancing, 2019 IEEE/CVF Conference on Computer Vision and Pattern Recognition (CVPR) (2019) 831--839.

\bibitem{Chaudhry2021UsingHT}
A.~Chaudhry, A.~Gordo, P.~Dokania, P.~H.~S. Torr, D.~Lopez-Paz, Using hindsight to anchor past knowledge in continual learning, in: AAAI, 2021.

\bibitem{Chaudhry2019EfficientLL}
A.~Chaudhry, M.~Ranzato, M.~Rohrbach, M.~Elhoseiny, Efficient lifelong learning with a-gem, ArXiv abs/1812.00420 (2019).

\bibitem{Buzzega2020DarkEF}
P.~Buzzega, M.~Boschini, A.~Porrello, D.~Abati, S.~Calderara, Dark experience for general continual learning: a strong, simple baseline, ArXiv abs/2004.07211 (2020).

\bibitem{Dam2022ScalableAO}
T.~Dam, M.~Pratama, M.~M. Ferdaus, S.~G. Anavatti, H.~Abbas, Scalable adversarial online continual learning, ArXiv abs/2209.01558 (2022).

\bibitem{VinciusdeCarvalho2022ClassIncrementalLV}
M.~V. de~Carvalho, M.~Pratama, J.~Zhang, Y.~San, Class-incremental learning via knowledge amalgamation, ArXiv abs/2209.02112 (2022).

\bibitem{Masum2023AssessorGuidedLF}
M.~A. Ma'sum, M.~Pratama, E.~D. Lughofer, W.~Ding, W.~Jatmiko, Assessor-guided learning for continual environments, Inf. Sci. 640 (2023) 119088.

\bibitem{Ven2019ThreeSF}
G.~M. van~de Ven, A.~Tolias, Three scenarios for continual learning, ArXiv abs/1904.07734 (2019).

\bibitem{Xue2022MetaattentionFV}
M.~Xue, H.~Zhang, J.~Song, M.~Song, Meta-attention for vit-backed continual learning, 2022 IEEE/CVF Conference on Computer Vision and Pattern Recognition (CVPR) (2022) 150--159.

\bibitem{Dosovitskiy2020AnII}
A.~Dosovitskiy, L.~Beyer, A.~Kolesnikov, D.~Weissenborn, X.~Zhai, T.~Unterthiner, M.~Dehghani, M.~Minderer, G.~Heigold, S.~Gelly, J.~Uszkoreit, N.~Houlsby, \href{https://api.semanticscholar.org/CorpusID:225039882}{An image is worth 16x16 words: Transformers for image recognition at scale}, ArXiv abs/2010.11929 (2020).
\newline\urlprefix\url{https://api.semanticscholar.org/CorpusID:225039882}

\bibitem{paeedeh2024crossdomain}
N.~Paeedeh, M.~Pratama, M.~A. Ma'sum, W.~Mayer, Z.~Cao, R.~Kowlczyk, Cross-domain few-shot learning via adaptive transformer networks (2024).
\newblock \href {http://arxiv.org/abs/2401.13987} {\path{arXiv:2401.13987}}.

\bibitem{Chi2022MetaFSCILAM}
Z.~Chi, L.~Gu, H.~Liu, Y.~Wang, Y.~Yu, J.~Tang, \href{https://api.semanticscholar.org/CorpusID:249979990}{Metafscil: A meta-learning approach for few-shot class incremental learning}, 2022 IEEE/CVF Conference on Computer Vision and Pattern Recognition (CVPR) (2022) 14146--14155.
\newline\urlprefix\url{https://api.semanticscholar.org/CorpusID:249979990}

\bibitem{Masum2023FewShotCL}
M.~A. Ma'sum, M.~Pratama, L.~Liu, E.~D. Lughofer, Habibullah, R.~Kowalczyk, Few-shot continual learning via flat-to-wide approaches, ArXiv abs/2306.14369 (2023).

\bibitem{pan2023ssfe}
Z.~Pan, X.~Yu, M.~Zhang, Y.~Gao, Ssfe-net: Self-supervised feature enhancement for ultra-fine-grained few-shot class incremental learning, in: Proceedings of the IEEE/CVF winter conference on applications of computer vision, 2023, pp. 6275--6284.

\bibitem{zhuang2023gkeal}
H.~Zhuang, Z.~Weng, R.~He, Z.~Lin, Z.~Zeng, Gkeal: Gaussian kernel embedded analytic learning for few-shot class incremental task, in: Proceedings of the IEEE/CVF Conference on Computer Vision and Pattern Recognition, 2023, pp. 7746--7755.

\end{thebibliography}






\end{document}